\documentclass{article}
\PassOptionsToPackage{numbers, sort}{natbib}

 \usepackage[preprint]{neurips_2025}

\usepackage{amssymb}
\newcommand{\inc}[1]{\textcolor{green!60!black}{$\uparrow$\small #1}}
\newcommand{\dec}[1]{\textcolor{red}{$\downarrow$\small #1}}

\usepackage[utf8]{inputenc} 
\usepackage[T1]{fontenc}    
\usepackage{hyperref}       
\usepackage{url}            
\usepackage{booktabs}       
\usepackage{amsfonts}       
\usepackage{wrapfig}
\usepackage{nicefrac}       
\usepackage{microtype}      
\usepackage{booktabs}
\usepackage{multirow}
\usepackage{graphicx}
\usepackage[table,xcdraw]{xcolor}
\usepackage{booktabs} 
\usepackage{caption}
\usepackage{subcaption}

\title{GEMS: Agent-Native Multimodal Generation with Memory and Skills}

\definecolor{color1}{HTML}{f2f3f5} 

\definecolor{color2}{HTML}{fff5e6}

\definecolor{color3}{HTML}{e0f7fa}   
\definecolor{color4}{HTML}{f2f2f2}

\author{%
  Zefeng He\textsuperscript{1,2}\thanks{Equal contribution. \quad $\dagger$ Corresponding authors.}  , 
  Siyuan Huang\textsuperscript{1,3*}, Xiaoye Qu\textsuperscript{1$\dagger$}, Yafu Li\textsuperscript{1,4}, 
  Tong Zhu\textsuperscript{1}, Yu Cheng\textsuperscript{4$\dagger$}, Yang Yang\textsuperscript{3$\dagger$} \\
  \textsuperscript{1}Shanghai AI Laboratory, \textsuperscript{2}Nanjing University, 
  \textsuperscript{3}Shanghai Jiao Tong University, \textsuperscript{4}CUHK
}

\begin{document}

\maketitle

\begin{center}
    \vspace{-25pt}
    \textbf{Project Page:} \href{https://gems-gen.github.io}{\textcolor{blue}{https://gems-gen.github.io}}
    \vspace{10pt}
\end{center}

\begin{abstract}

Recent multimodal generation models have achieved remarkable progress on general-purpose generation tasks, yet continue to struggle with complex instructions and specialized downstream tasks. Inspired by the success of advanced agent frameworks such as Claude Code, we propose \textbf{GEMS} (Agent-Native Multimodal \textbf{GE}neration with \textbf{M}emory and \textbf{S}kills), a framework that pushes beyond the inherent limitations of foundational models on both general and downstream tasks. GEMS is built upon three core components. Agent Loop introduces a structured multi-agent framework that iteratively improves generation quality through closed-loop optimization. Agent Memory provides a persistent, trajectory-level memory that hierarchically stores both factual states and compressed experiential summaries, enabling a global view of the optimization process while reducing redundancy. Agent Skill offers an extensible collection of domain-specific expertise with on-demand loading, allowing the system to effectively handle diverse downstream applications. Across five mainstream tasks and four downstream tasks, evaluated on multiple generative backends, GEMS consistently achieves significant performance gains. Most notably, it enables the lightweight 6B model Z-Image-Turbo to surpass the state-of-the-art Nano Banana 2 on
GenEval2, demonstrating the effectiveness of agent harness in extending model capabilities beyond their original limits.
\end{abstract}    

\begin{center}
    \centering
    \includegraphics[width=\textwidth]{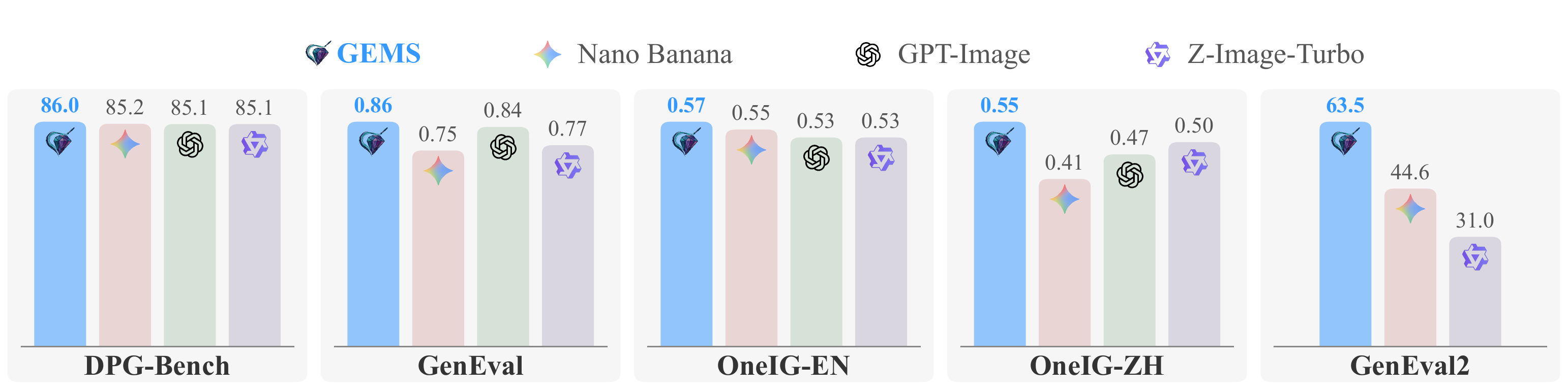}
\captionof{figure}{{Overall Performance of GEMS with Z-Image-Turbo.} GEMS enables a lightweight, distilled 6B model Z-Image-Turbo to outperform prominent closed-source models, such as Nano Banana and GPT-Image, across various mainstream benchmarks.}
    \label{fig:teaser}
\end{center}

\section{Introduction}

Multimodal generation has undergone transformative growth in recent years~\cite{ramesh2022hierarchical,saharia2022photorealistic,brooks2024video}, where advanced algorithms~\cite{ho2020denoising,song2020denoising,ho2022classifier,lipman2022flow,liu2022flow,albergo2022building,chen2025flash} and architectural designs~\cite{rombach2022high,peebles2023scalable,mmdit} have significantly enhanced the quality and accessibility of visual synthesis. Leading closed-source models, such as GPT-Image and Nano Banana, alongside prominent open-source frameworks like Qwen-Image~\cite{qwen-image} and Z-Image~\cite{z-image}, have set new state-of-the-art records across various benchmarks. These models exhibit remarkable proficiency in handling mainstream and straightforward tasks~\cite{geneval,dpg,oneig}, consistently producing high-fidelity results that align closely with general-purpose textual prompts. Despite these achievements, they often struggle when handling intricate, multi-faceted instructions~\cite{geneval2} or specialized downstream applications~\cite{crea,spatialgeneval,longtextbench}, which constitutes the persistent ``long-tail'' challenge where general-purpose capabilities reach their limits.

To bridge these gaps, inference-time scaling~\cite{promptist,ma2025inference} has emerged as a pivotal strategy for enhancing model performance. Current research primarily focuses on iterative refinement loops~\cite{jiang2026genagent,li2025editthinker,guo2025thinking} or multi-agent collaborative systems~\cite{li2025mccd,maestro,craft} to tackle complex tasks. Meanwhile, specialized multi-agent frameworks have been developed for targeted downstream domains, such as creative drawing~\cite{crea} and academic illustration~\cite{zhu2026paperbanana}, to provide domain-specific optimizations.
However, existing multi-agent systems face several critical limitations. Frameworks such as Maestro~\cite{maestro} rely on successive single-step updates, while many iterative approaches~\cite{jiang2026genagent,li2025editthinker} simply accumulate historical context, leading to either insufficient guidance or excessive information redundancy. On the other hand, while systems optimized for specific downstream tasks~\cite{spatialgeneval,crea,zhu2026paperbanana} achieve localized success, they are often difficult to integrate with mainstream generative pipelines due to their specialized coordination mechanisms, resulting in fragmented and less adaptable architectures.

Inspired by recent breakthroughs in pioneering agent frameworks such as Claude Code and OpenClaw, we propose \textbf{GEMS} (Agent-Native Multimodal \textbf{GE}neration with \textbf{M}emory and \textbf{S}kills), a framework redesigned from an innovative agentic perspective. GEMS is specifically architected to overcome the limitations in complex instructions and specialized downstream tasks through three core pillars:
(1) \textbf{Agent Loop}, which introduces a structured multi-agent framework that iteratively improves generation quality through closed-loop optimization, thereby ensuring high-fidelity performance on complex tasks~\cite{geneval2}; 
(2) \textbf{Agent Memory}, a persistent mechanism that, unlike simple context accumulation~\cite{jiang2026genagent} or successive single-step updates~\cite{maestro}, maintains a global record of the optimization trajectory while utilizing hierarchical compression to preserve factual artifacts while distilling high-level experiences, effectively eliminating information redundancy and improving the overall quality of iterative refinement; 
(3) \textbf{Agent Skill}, an extensible repository of domain-specific expertise that resolves the fragmentation of isolated task-specific systems~\cite{crea,zhu2026paperbanana} by employing an on-demand loading and progressive exposure mechanism to maximize scalability and minimize cognitive load, allowing the system to effectively handle diverse downstream tasks.
By integrating these components, GEMS transcends the constraints of traditional iterative loops, offering a more scalable and intelligent solution for complex instructions and downstream tasks.

To validate the effectiveness of GEMS, we conducted extensive experiments across nine distinct tasks, including five challenging mainstream benchmarks such as GenEval2~\cite{geneval2} and four specialized downstream tasks spanning diverse domains. Our framework's generalizability was verified across multiple generative backends. Specifically, leveraging the lightweight, distilled Z-Image-Turbo~\cite{z-image}, GEMS yielded significant average performance gains of 14.22 on mainstream benchmarks and 14.03 on downstream tasks. Most notably, our framework enables the 6B Z-Image-Turbo to surpass the state-of-the-art Nano Banana 2 on GenEval2, demonstrating that agentic reasoning and domain-specific expertise can effectively push beyond the inherent boundaries of foundational models. We further validated our framework on another mainstream open-source model, Qwen-Image-2512~\cite{qwen-image}, where it achieved average improvements of 16.24 and 7.96 across mainstream and downstream tasks, respectively. These results underscore the robust generalizability and scalability of our agentic system across varying model architectures and scales.

In summary, our primary contributions are as follows:

\begin{itemize}
    \item We propose \textbf{GEMS}, an agent-native multimodal generation framework that employs iterative refinement to significantly enhance performance in complex generation tasks.
    \item We introduce a persistent \textbf{Agent Memory} mechanism utilizing hierarchical compression, which efficiently manages historical context in multi-turn optimization trajectories.
    \item We develop an extensible \textbf{Agent Skill} module utilizing efficient on-demand loading to equip the system with domain-specific expertise for specialized downstream applications.
    \item Extensive experiments across nine diverse tasks validate the effectiveness of GEMS, highlighting the transformative potential of agentic frameworks for multimodal generation.
\end{itemize}

\section{Related Works}
\subsection{Inference-Time Scaling for Multimodal Generation}
Recent years have witnessed significant progress in multimodal generation~\cite{qwen-image,z-image,brooks2024video,he2025diffthinker,saharia2022photorealistic,ramesh2022hierarchical}, and inference-time scaling has emerged as a promising strategy for performance enhancement. Early approaches primarily relied on simple prompt rewriting~\cite{promptist} or random search~\cite{ma2025inference} to optimize generation. Other methods~\cite{fang2025got,jiang2025t2i,liao2025imagegen,jiao2025thinkgen,kou2026think} introduced Chain-of-Thought (CoT)~\cite{wei2022chain} reasoning to provide more guidance for multimodal generation. More advanced approaches~\cite{wang2025imagent,jiang2026genagent,mondal2025countloop,li2025reflect,zhuo2025reflection,li2025editthinker,guo2025thinking} have adopted iterative refinement loops to progressively optimize the results. 
Recent studies~\cite{li2025mccd,park2026guiding,maestro,craft,crea,zhu2026paperbanana} have also explored multi-agent systems. Some approaches~\cite{maestro,craft} leverage multi-agent collaboration and iterative optimization to enhance the generation process in complex tasks, yet are still limited to basic agent loops. Other  studies~\cite{crea,zhu2026paperbanana} focus on customized designs for specific downstream tasks, yet are often difficult to integrate with mainstream generative workflows. In contrast, GEMS adopts advanced agentic paradigms to address these limitations.

\subsection{Agent Systems}
Agent systems serve as autonomous frameworks that extend the reasoning and execution capabilities of LLMs through structured planning and interaction. Foundational works established agent loops~\cite{yao2022react,shinn2023reflexion,wang2023plan,madaan2023self,schick2023toolformer,he2025framethinker} that enable models to alternate between reasoning and acting within a self-correcting cycle. Building on this, multi-agent systems~\cite{li2023camel,hong2023metagpt,wu2024autogen,chen2023agentverse} employ specialized roles that collaborate through communication protocols to tackle more intricate objectives. Furthermore, the integration of agent memory~\cite{packer2023memgpt,chhikara2025mem0,xu2025mem,long2025seeing,fu2026latentmem} enhances system performance in long-context and multi-turn interactions. More recently, agent skills~\cite{xia2026skillrl,liang2026skillnet,li2026skillsbench,jiang2026xskill} have further expanded the boundaries of agent systems, empowering them to execute complex tasks through domain-specific workflows. Building upon these capabilities, state-of-the-art agent systems such as Claude Code and OpenClaw have demonstrated remarkable capabilities in executing sophisticated, real-world operations, inspiring our adaptation of these agentic paradigms to multimodal generation.
\section{Method}
\begin{figure}[t]
\centering
\includegraphics[width=\textwidth]{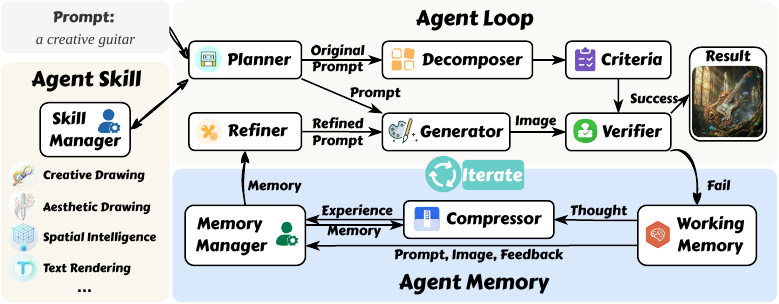}
\caption{{The system architecture of GEMS.} The framework consists of three primary pillars: Agent Loop, Agent Memory, and Agent Skill. The user prompt is augmented with domain-specific expertise from Agent Skill, and then iteratively refined within the Agent Loop, with Agent Memory managing the historical context to guide the generation process.}
\label{fig:flowchart}
\end{figure}

As shown in Figure~\ref{fig:flowchart}, GEMS comprises three core components: Agent Loop, Agent Memory, and Agent Skill. These modules collaborate to address the challenges of complex instruction following and specialized downstream tasks. The following subsections describe each component in detail.

\subsection{Agent Loop}
 Agent Loop serves as the backbone of GEMS, comprising several collaborative modules: Planner, Decomposer, Generator, Verifier, and Refiner.
 
\noindent \textbf{Planner.} The Planner, denoted as $\mathcal{F}_{plan}$, serves as the strategic entry point of the system. It first interacts with the Skill Manager to identify relevant expertise from the domain-specific repository $\mathcal{S}$ (Sec. {3.3}) based on the user prompt $U$. This interaction retrieves a subset of triggered skills $\mathcal{S}_{trig} \subseteq \mathcal{S}$; if the task does not align with any specialized domain, $\mathcal{S}_{trig}$ remains empty. Leveraging the retrieved skills (if any), the Planner synthesizes an enhanced initial prompt $P_1$ designed to provide superior guidance for the generation process. Concurrently, it dispatches the original prompt $U$ to the Decomposer to establish the foundational evaluation framework. The operation is defined as:
\begin{equation}
    (P_1, U) = \mathcal{F}_{plan}(U, \mathcal{S}).
\end{equation}

\noindent \textbf{Decomposer.} To ensure fine-grained evaluation, the Decomposer $\mathcal{F}_{dec}$ partitions the user's original prompt $U$ into a set of atomic visual requirements $\mathcal{C} = \{c_1, c_2, \dots, c_n\}$. Each criterion $c_j$ is formulated as a binary (yes/no) probe that represents an essential semantic or structural constraint:
\begin{equation}
    \mathcal{C} = \mathcal{F}_{dec}(U).
\end{equation}

\noindent \textbf{Generator.} The Generator $\mathcal{F}_{gen}$ is a model-agnostic module responsible for synthesizing the visual output. At each iteration $i$, it produces an image $I_i$ based on the current optimized prompt $P_i$:
\begin{equation}
    I_i = \mathcal{F}_{gen}(P_i).
\end{equation}

\noindent \textbf{Verifier.} The Verifier $\mathcal{F}_{ver}$, powered by a Multimodal Large Language Model (MLLM), assesses the generated image $I_i$ against the predefined atomic criteria set $\mathcal{C}$. It maps the visual and textual inputs to a binary feedback vector $V_i = \{v_{i,1}, \dots, v_{i,n}\}$:
\begin{equation}
    V_i = \mathcal{F}_{ver}(I_i, \mathcal{C}), \quad v_{i,j} \in \{0, 1\}.
\end{equation}
The system then executes a conditional branch based on the result of $V_i$. If all criteria are met (i.e., $\forall j, v_{i,j} = 1$), the iterative loop terminates, and $I_i$ is returned as the final output. If any criterion remains unsatisfied and the current iteration $i$ is below the maximum limit $N_{max}$, the vector $V_i$ is dispatched to the Refiner as diagnostic feedback. However, should the system reach $N_{max}$ without satisfying all criteria, it performs a global evaluation over the optimization trajectory and returns the image $I_{best}$ that fulfilled the maximum number of requirements:
\begin{equation}
    I_{best} = \arg\max_{I_k} \sum_{j=1}^{n} v_{k,j}, \quad k \in \{1, \dots, N_{max}\}.
\end{equation}

\noindent \textbf{Refiner.} The Refiner $\mathcal{F}_{ref}$ facilitates prompt evolution by closing the feedback loop. At iteration $i$, it synthesizes a refined prompt $P_{i+1}$ by analyzing the current state and historical context. Crucially, $\mathcal{M}_{i-1}$ represents the state of the Agent Memory at the conclusion of iteration $i-1$, which encapsulates the cumulative trajectory of preceding attempts. The Refiner integrates the current prompt $P_i$, the generated image $I_i$, the verification feedback $V_i$, and the internal reasoning $T_i$ (reflecting the MLLM's thought process during refinement) with this historical state $\mathcal{M}_{i-1}$ to derive the next-turn prompt:
\begin{equation}
    P_{i+1} = \mathcal{F}_{ref}(P_i, I_i, V_i, T_i, \mathcal{M}_{i-1}).
\end{equation}

\subsection{Agent Memory}

Previous multimodal agent systems, such as Maestro~\cite{maestro}, often adopt an evolutionary design that only focuses on the immediate previous result or the best-performing state, lacking a comprehensive historical perspective across the entire generation process. 
To transcend the limitations of simple successive single-step updates, we implement a persistent memory mechanism that maintains a global record of the optimization trajectory. To optimize for both information density and token efficiency, we propose a Hierarchical Compression strategy to manage the historical context.
Specifically, we categorize the iteration state into two distinct tiers. Factual artifacts with minimal token footprints, such as the prompt $P_i$, the generated image $I_i$, and the verification feedback $V_i$, serve as reliable and objective data points and are archived in their raw form to ensure historical accuracy. Conversely, reasoning traces $T_i$, which are often verbose and redundant~\cite{qu2025survey,sui2025stopoverthinkingsurveyefficient}, are processed by a \textbf{Compressor} $\mathcal{F}_{comp}$ to distill them into concise, high-level experiences $E_i$:
\begin{equation}
    E_i = \mathcal{F}_{comp}(P_i, I_i, V_i, T_i, \mathcal{M}_{i-1}).
\end{equation}
The resulting memory state $\mathcal{M}_i$ is then updated as a sequence of these hybrid state tuples, ensuring that the system retains both factual anchors and strategic reflections:
\begin{equation}
    \mathcal{M}_i = \{(P_1, I_1, V_1, E_1), \dots, (P_i, I_i, V_i, E_i)\}.
\end{equation}
By archiving this hierarchically compressed representation, the system eliminates informational noise while providing the Refiner with a robust, long-context perspective of the entire generation trajectory.

\subsection{Agent Skill}

Conventional agent systems often rely on task-specific implementations~\cite{crea,zhu2026paperbanana} for downstream applications; however, these specialized designs are difficult to integrate with mainstream generative pipelines, resulting in fragmented and less adaptable architectures. To address these limitations and enhance downstream performance, we introduce the Agent Skill module, a repository of domain-specific expertise that allows the system to transcend general-purpose limitations. The Planner interacts with this module at the initial stage of the pipeline, matching user intent with specialized skills to obtain an enhanced prompt before the iterative loop begins.

\begin{wrapfigure}{r}{0.5\columnwidth} 
    \centering
    \vspace{-15pt} 
    \includegraphics[width=0.48\columnwidth]{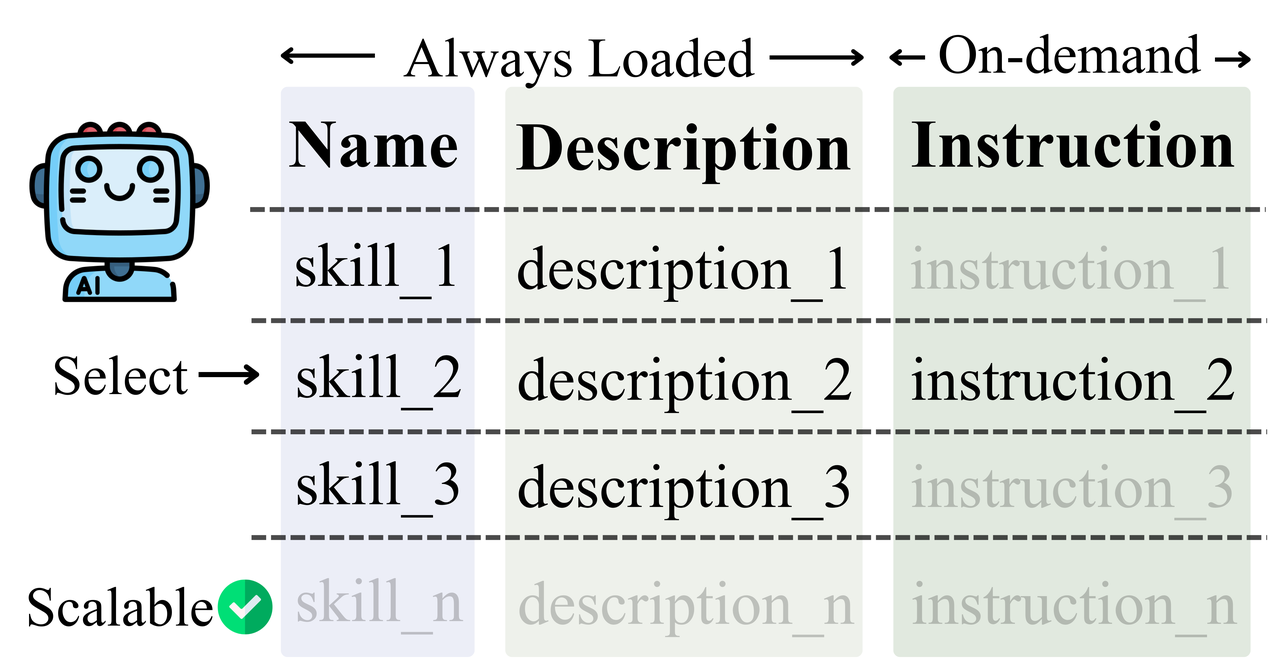}
    \caption{Architecture of the Agent Skill system, highlighting its scalable and on-demand nature.}
    \label{fig:skill}
    \vspace{-10pt} 
\end{wrapfigure}

As illustrated in Figure~\ref{fig:skill}, our system features an on-demand loading and progressive exposure mechanism. To ensure token efficiency, only the names and descriptions of skills are ``always loaded'' as a lightweight manifest. The comprehensive instructions, which contain dense domain knowledge, are fetched only when a specific skill is triggered. 
This design directly enables high scalability and user-friendliness. Because detailed instructions are loaded only when necessary, the system can support an extensive library of expertise without imposing a significant computational or cognitive burden on the reasoning process. Furthermore, it minimizes the barrier for contributors; users are not required to understand the full operational logic of the system. By simply providing a markdown file (e.g., \texttt{SKILL.md}) that outlines the relevant information, the system can automatically understand and activate the new skill, empowering users to generate any content with significantly enhanced fidelity and domain-specific precision. Such modularity ensures the system remains accessible and adaptable to increasingly diverse requirements.

\begin{table*}[t]
\centering
\caption{Evaluation on Mainstream Tasks. For inference-time scaling methods, all results are evaluated by us, with the best and second-best performances highlighted in \textbf{bold} and \underline{underlined}, and relative improvements or decreases compared to the baseline indicated by \inc{} and \dec{}, respectively. Other results are sourced from public data. The ``Avg." column represents the mean of normalized scores, with OneIG-EN and OneIG-ZH pre-averaged as a single metric before final aggregation.}
\label{tab:mainstream_results}
\resizebox{\textwidth}{!}{
\begin{tabular}{l|l|cccccc|c}
\toprule
\textbf{Model} & \textbf{Method} & \textbf{GenEval} & \textbf{GenEval2} & \textbf{DPG-Bench} & \textbf{OneIG-EN} & \textbf{OneIG-ZH} & \textbf{WISE} & \textbf{Avg.} \\ \midrule

\rowcolor{color1} \multicolumn{9}{c}{\textbf{Closed-Source Models}} \\ \midrule
Nano Banana  & -- & 0.75 & 44.6 & 85.23 & 0.550 & 0.405 & -- & -- \\
GPT-Image 1  & -- & 0.84 & -- & 85.15 & 0.533 & 0.474 & 0.80 & -- \\
Seedream 4~\cite{seedream4}  & -- & 0.84 & -- & 88.25 & 0.576 & 0.553 & 0.78 & -- \\
\midrule

\rowcolor{color2} \multicolumn{9}{c}{\textbf{Open-Source Models}} \\ \midrule
Bagel~\cite{bagel}  & -- & 0.82 & 23.1 & 85.10 & 0.361 & 0.370 & 0.70 & 59.35 \\
Z-Image~\cite{z-image}  & -- & 0.84 & -- & 88.14 & 0.546 & 0.535 & -- & -- \\
Qwen-Image~\cite{qwen-image}  & -- & 0.87 & 33.8 & 88.32 & 0.539 & 0.548 & 0.62 & 65.09 \\
\midrule

\rowcolor{color3} \multicolumn{9}{c}{\textbf{Inference-Time Scaling}} \\ \midrule
\multirow{7}{*}{Z-Image-Turbo~\cite{z-image}} & Original & 0.77 \inc{0.00} & 31.0 \inc{0.0} & 85.08 \inc{0.00} & 0.526 \inc{0.000} & 0.501 \inc{0.000} & 0.57 \inc{0.00} & 60.29 \\
 & Rewrite & 0.77 \inc{0.00} & 39.7 \inc{8.7} & 84.48 \dec{0.60} & 0.520 \dec{0.006} & 0.469 \dec{0.032} & \underline{0.84} \inc{0.27} & 66.93 \\
 & Promptist~\cite{promptist} & 0.76 \dec{0.01} & 32.3 \inc{1.3} & 65.69 \dec{19.39} & 0.333 \dec{0.193} & 0.327 \dec{0.174} & 0.54 \dec{0.03} & 52.20 \\
 & Search~\cite{ma2025inference} & 0.81 \inc{0.04} & 43.5 \inc{12.5} & \underline{85.47} \inc{0.39} & 0.530 \inc{0.004} & 0.506 \inc{0.005} & 0.61 \inc{0.04} & 64.55 \\
 & Maestro~\cite{maestro} & \underline{0.82} \inc{0.05} & 44.6 \inc{13.6} & 85.29 \inc{0.21} & 0.548 \inc{0.022} & 0.519 \inc{0.018} & \textbf{0.85} \inc{0.28} & 70.05 \\
 & CRAFT~\cite{craft} & 0.80 \inc{0.03} & \underline{62.4} \inc{31.4} & 85.29 \inc{0.21} & \textbf{0.582} \inc{0.056} & \underline{0.542} \inc{0.041} & 0.78 \inc{0.21} & \underline{72.38} \\
 \rowcolor{color4} \cellcolor{white} & GEMS (Ours)& \textbf{0.86} \inc{0.09} & \textbf{63.5} \inc{32.5} & \textbf{86.01} \inc{0.93} & \underline{0.569} \inc{0.043} & \textbf{0.552} \inc{0.051} & 0.81 \inc{0.24} & \textbf{74.51} \\
 \midrule
\multirow{7}{*}{Qwen-Image-2512~\cite{qwen-image}} & Original & 0.66 \inc{0.00} & 29.0 \inc{0.0} & 84.69 \inc{0.00} & 0.487 \inc{0.000} & 0.489 \inc{0.000} & 0.59 \inc{0.00} & 57.50 \\
 & Rewrite & 0.68 \inc{0.02} & 39.8 \inc{10.8} & 83.49 \dec{1.20} & 0.465 \dec{0.022} & 0.435 \dec{0.054} & \underline{0.81} \inc{0.22} & 63.46 \\
 & Promptist~\cite{promptist} & 0.66 \inc{0.00} & 27.4 \dec{1.6} & 64.02 \dec{20.67} & 0.288 \dec{0.199} & 0.307 \dec{0.182} & 0.57 \dec{0.02} & 48.83 \\
 & Search~\cite{ma2025inference} & 0.74 \inc{0.08} & 42.1 \inc{13.1} & \textbf{86.17} \inc{1.48} & 0.502 \inc{0.015} & 0.497 \inc{0.008} & 0.67 \inc{0.08} & 63.84 \\
 & Maestro~\cite{maestro} & 0.76 \inc{0.10} & 48.0 \inc{19.0} & 84.07 \dec{0.62} & 0.494 \inc{0.007} & 0.489 \inc{0.000} & \textbf{0.84} \inc{0.25} & 68.24 \\
 & CRAFT~\cite{craft} & \textbf{0.79} \inc{0.13} & \underline{66.3} \inc{37.3} & \underline{85.87} \inc{1.18} & \underline{0.533} \inc{0.046} & \underline{0.518} \inc{0.029} & 0.79 \inc{0.20} & \underline{72.54} \\
 \rowcolor{color4} \cellcolor{white} & GEMS (Ours) & \textbf{0.79} \inc{0.13} & \textbf{70.4} \inc{41.4} & 85.59 \inc{0.90} & \textbf{0.542} \inc{0.055} & \textbf{0.532} \inc{0.043} & 0.80 \inc{0.21} & \textbf{73.74} \\
 \bottomrule
\end{tabular}
}
\end{table*}

\begin{table*}[t]
\centering
\caption{Evaluation on Downstream Tasks. For inference-time scaling methods, all results are evaluated by us, with the best and second-best performances highlighted in \textbf{bold} and \underline{underlined}, and relative improvements or decreases compared to the baseline indicated by \inc{} and \dec{}, respectively. Other results are sourced from public data. The ``Avg.'' column represents the mean of normalized scores, with LongText-EN and LongText-ZH pre-averaged as a single metric before final aggregation.}
\label{tab:downstream_results}
\resizebox{\textwidth}{!}{
\begin{tabular}{l|l|ccccc|c}
\toprule
\textbf{Model} & \textbf{Method} & \textbf{LongText-EN} & \textbf{LongText-ZH} & \textbf{SpatialGenEval} & \textbf{CREA} & \textbf{ArtiMuse} & \textbf{Avg.} \\ \midrule

\rowcolor{color1} \multicolumn{8}{c}{\textbf{Closed-Source Models}} \\ \midrule
Nano Banana  & -- & -- & -- & 61.7 & -- &-- & --\\
GPT-Image 1  & -- & 0.956 & 0.619 & 60.5 & -- &-- & --\\
Seedream 4~\cite{seedream4}  & -- & -- & -- & 62.7 & -- & --& --\\
\midrule

\rowcolor{color2} \multicolumn{8}{c}{\textbf{Open-Source Models}} \\ \midrule
Bagel~\cite{bagel}  & -- & 0.373 & 0.310 & 57.0 & -- & --& --\\
Z-Image~\cite{z-image}  & -- & 0.935 & 0.936 & -- & -- &-- & -- \\
Qwen-Image~\cite{qwen-image}  & -- & 0.943 & 0.946 & 60.6 & -- & --& -- \\
\midrule

\rowcolor{color3} \multicolumn{8}{c}{\textbf{Inference-Time Scaling}} \\ \midrule
\multirow{7}{*}{Z-Image-Turbo~\cite{z-image}} & Original & 0.912 \inc{0.000} & 0.932 \inc{0.000} & 58.7 \inc{0.0} & 11.84 \inc{0.00} & 43.27 \inc{0.00} & 58.41 \\
 & Rewrite& 0.571 \dec{0.341} & 0.495 \dec{0.437} & 59.6 \inc{0.9} & \underline{16.75} \inc{4.91} & \underline{57.86} \inc{14.59} & 56.65 \\
 & Promptist~\cite{promptist} & 0.034 \dec{0.878} & 0.070 \dec{0.862} & 48.3 \dec{10.4} & 14.00 \inc{2.16} & 43.74 \inc{0.47} & 35.98 \\
 & Search~\cite{ma2025inference} & 0.918 \inc{0.006} & \underline{0.937} \inc{0.005} & 59.3 \inc{0.6} & 12.98 \inc{1.14} & 44.21 \inc{0.94} & 59.88 \\
 & Maestro~\cite{maestro} & 0.877 \dec{0.035} & 0.807 \dec{0.125} & 60.3 \inc{1.6} & 15.81 \inc{3.97} & 56.86 \inc{13.59} & \underline{63.52} \\
 & CRAFT~\cite{craft} & \underline{0.951} \inc{0.039} & 0.760 \dec{0.172} & \underline{60.6} \inc{1.9} & 13.63 \inc{1.79} & 54.95 \inc{11.68} & 61.63 \\
 \rowcolor{color4} \cellcolor{white}&GEMS (Ours)& \textbf{0.952} \inc{0.040} & \textbf{0.940} \inc{0.008} & \textbf{61.4} \inc{2.7} & \textbf{22.55} \inc{10.71} & \textbf{58.58} \inc{15.31} & \textbf{72.44} \\

 \midrule
\multirow{7}{*}{Qwen-Image-2512~\cite{qwen-image}} & Original & 0.854 \inc{0.000} & 0.892 \inc{0.000} & 60.0 \inc{0.0} & 17.45 \inc{0.00} & 59.94 \inc{0.00} & 66.35 \\
 & Rewrite& 0.563 \dec{0.291} & 0.468 \dec{0.424} & 60.3 \inc{0.3} & 18.12 \inc{0.67} & \underline{62.23} \inc{2.29} & 58.62 \\
 & Promptist~\cite{promptist} & 0.030 \dec{0.824} & 0.058 \dec{0.834} & 48.7 \dec{11.3} & 18.35 \inc{0.90} & 60.65 \inc{0.71} & 43.73 \\
 & Search~\cite{ma2025inference} & 0.883 \inc{0.029} & \underline{0.907} \inc{0.015} & 60.6 \inc{0.6} & \underline{18.71} \inc{1.26} & 59.96 \inc{0.02} & \underline{68.11} \\
 & Maestro~\cite{maestro} & 0.833 \dec{0.021} & 0.792 \dec{0.100} & 61.0 \inc{1.0} & 17.78 \inc{0.33} & 61.62 \inc{1.68} & 65.78 \\
 & CRAFT~\cite{craft} & \underline{0.897} \inc{0.043} & 0.787 \dec{0.105} & \underline{61.6} \inc{1.6} & 15.61 \dec{1.84} & 59.78 \dec{0.16} & 64.40 \\
 \rowcolor{color4} \cellcolor{white}&GEMS (Ours)& \textbf{0.913} \inc{0.059} & \textbf{0.931} \inc{0.039} & \textbf{62.1} \inc{2.1} & \textbf{24.01} \inc{6.56} & \textbf{62.89} \inc{2.95} & \textbf{74.31} \\
 \bottomrule
\end{tabular}
}
\end{table*}

\section{Experiments}
\subsection{Experimental Setup}
\paragraph{Implementation Details.}
To evaluate the effectiveness of GEMS, we conduct experiments with two distinct generative models: 
(1) \textbf{Z-Image-Turbo~\cite{z-image}}: Z-Image is an efficient 6B model, and we further utilize its distilled version, Z-Image-Turbo, to prioritize inference efficiency. 
(2) \textbf{Qwen-Image-2512~\cite{qwen-image}}\footnote{Our evaluations indicate that Qwen-Image-2512 exhibits lower benchmark scores than Qwen-Image. This finding is consistent with results reported in other recent studies, such as the GLM-Image Technical Blog~\cite{glm-image}.}: A representative 20B open-source model employed to verify the effectiveness of GEMS across different model architectures and parameter scales. We utilize Kimi K2.5~\cite{kimi2.5} as the MLLM backend.
By default, the maximum number of iterations is set to 5. Four skills tailored to our evaluation tasks are enabled: Creative Drawing, Aesthetic Drawing, Text Rendering, and Spatial Intelligence. Max number of triggered skills is set to 1, aligning with the singular focuses of the evaluation tasks. Further details are provided in Appendix~\ref{sec:eval}.

\paragraph{Benchmarks and Baselines}
We evaluate our system across five mainstream benchmarks, including GenEval~\cite{geneval}, GenEval2~\cite{geneval2}, DPG-Bench~\cite{dpg}, OneIG~\cite{oneig}, and WISE~\cite{wise}, and further incorporate LongText-Bench~\cite{longtextbench}, SpatialGenEval~\cite{spatialgeneval}, CREA~\cite{crea}, and ArtiMuse~\cite{cao2025artimuse}, as downstream tasks. Our baselines consist of strong closed-source and open-source generative models, as well as inference-time scaling systems. To ensure a fair comparison under similar computational budgets, we specifically set the parallelism factor for Search~\cite{ma2025inference} to 5, and limit the maximum number of iterations for Maestro~\cite{maestro} and CRAFT~\cite{craft} to 3 and 5, respectively. A detailed comparison of the computational costs versus performance gains for the various methods is illustrated in Figure~\ref{fig:trade_off}. Further details regarding the tasks and baselines are provided in Appendix~\ref{sec:benchmark} and Appendix~\ref{sec:baseline}.

\subsection{Main Results}

Tables~\ref{tab:mainstream_results} and~\ref{tab:downstream_results} present the experimental results across mainstream and downstream tasks, respectively. On mainstream tasks, GEMS, leveraging Z-Image-Turbo, achieves consistent performance gains with an average increase of 14.22 in normalized scores, outperforming prior inference-time scaling baselines. Further validation on Qwen-Image-2512 confirms the generalizability and effectiveness of our approach across different foundational architectures.

On downstream tasks, GEMS demonstrates an even more pronounced advantage, yielding an average improvement of 14.03 in normalized scores with Z-Image-Turbo, and significantly surpassing the best-performing inference-time scaling baseline (+8.92). Notably, we observe that several baseline methods involving prompt rewriting, such as Rewrite and Promptist~\cite{promptist}, exhibit significant performance degradation in certain tasks, particularly in text rendering. This decline stems from the fact that general-purpose rewriting strategies often lack domain-specific constraints, frequently compromising strict textual information during the optimization process. In contrast, GEMS incorporates specialized skills to provide targeted guidance for optimization, resulting in consistent and substantial performance enhancements even in highly specialized domains.

\subsection{Ablation and Discussion}

\begin{figure}[htbp]
    \vspace{-10pt}
    \centering
    \begin{minipage}{0.49\textwidth}
        \centering
        \includegraphics[width=\linewidth]{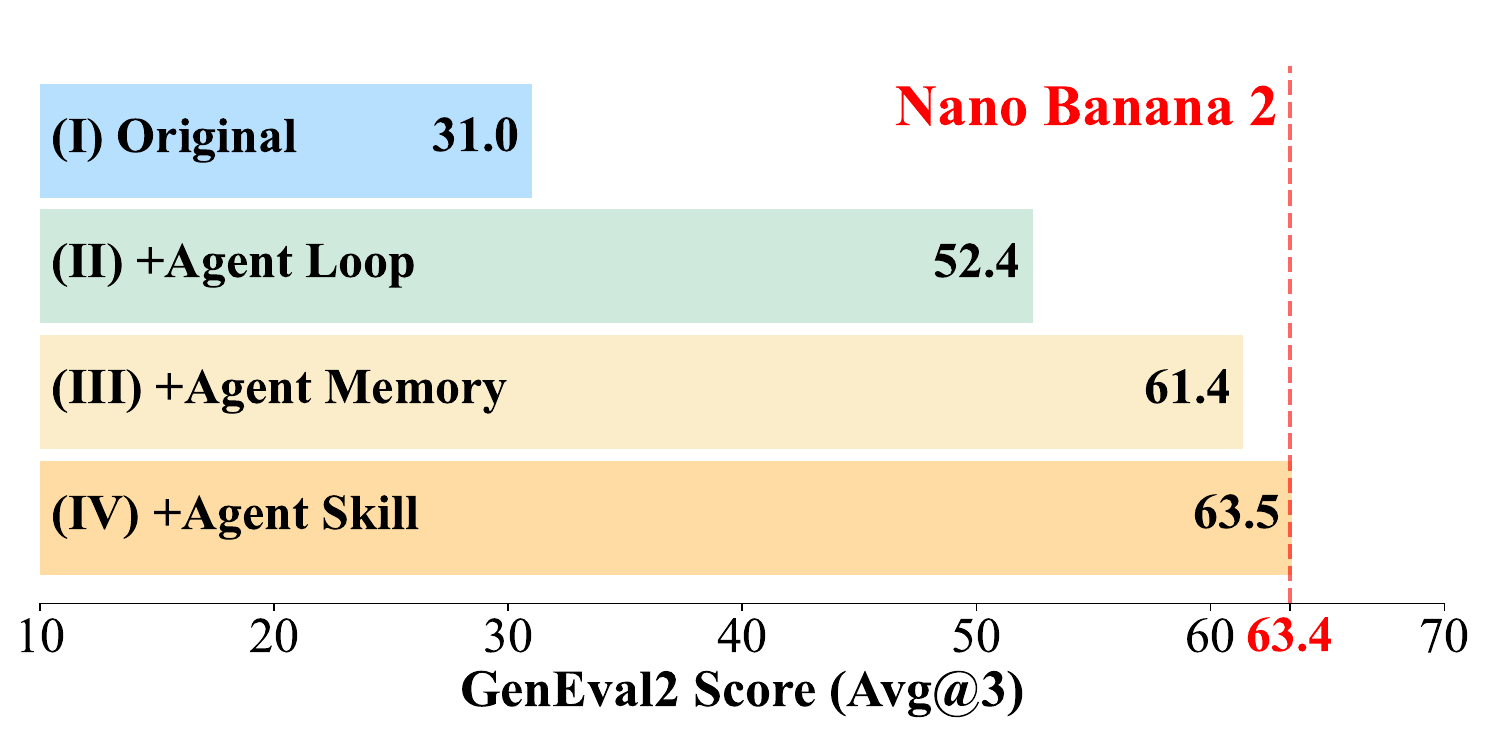}
    \end{minipage}
    \hfill
    \begin{minipage}{0.49\textwidth}
        \centering
        \includegraphics[width=\linewidth]{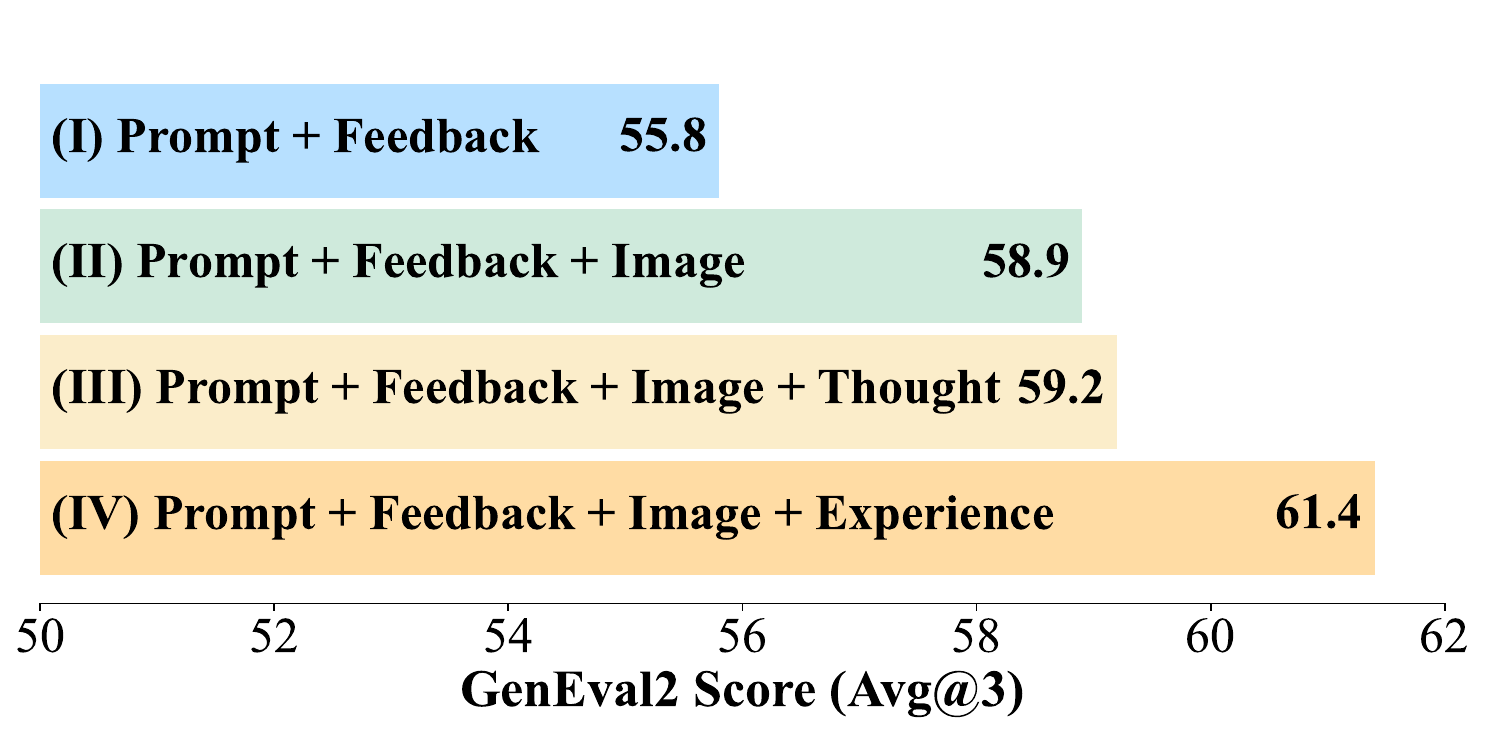}
    \end{minipage}
    \caption{Ablation study on GenEval2 with Z-Image-Turbo. (Left) Performance gains contributed by individual components, including Agent Loop, Agent Memory, and Agent Skill. (Right) Detailed analysis of the performance improvements of Agent Memory.}
    \label{fig:ablation_results}
    \vspace{-10pt}
\end{figure}

\paragraph{Overall Ablation Study.}
We ablate GEMS on GenEval2, selected due to its status as a challenging and unsaturated benchmark. To ensure robustness, we report results averaged over three independent runs using Z-Image-Turbo. As shown in Figure~\ref{fig:ablation_results}(left), the sequential integration of the Agent Loop, Agent Memory, and Agent Skill yields substantial performance gains. Specifically, the basic Agent Loop improves the score from 31.0 to 52.4, while the addition of Agent Memory and Agent Skill contributes further increases of 9.0 and 2.1 points, respectively, culminating in a final score of 63.5. Notably, GEMS enables the lightweight generator to outperform the state-of-the-art Nano Banana 2. This demonstrates that GEMS effectively unlocks the potential of foundational models, allowing them to transcend inherent capacity limits through agentic reasoning and domain-specific expertise.

\begin{wrapfigure}{r}{0.5\columnwidth}
    \centering
    \vspace{-18pt} 
    \begin{subfigure}{0.49\linewidth}
        \centering
        \includegraphics[width=\linewidth]{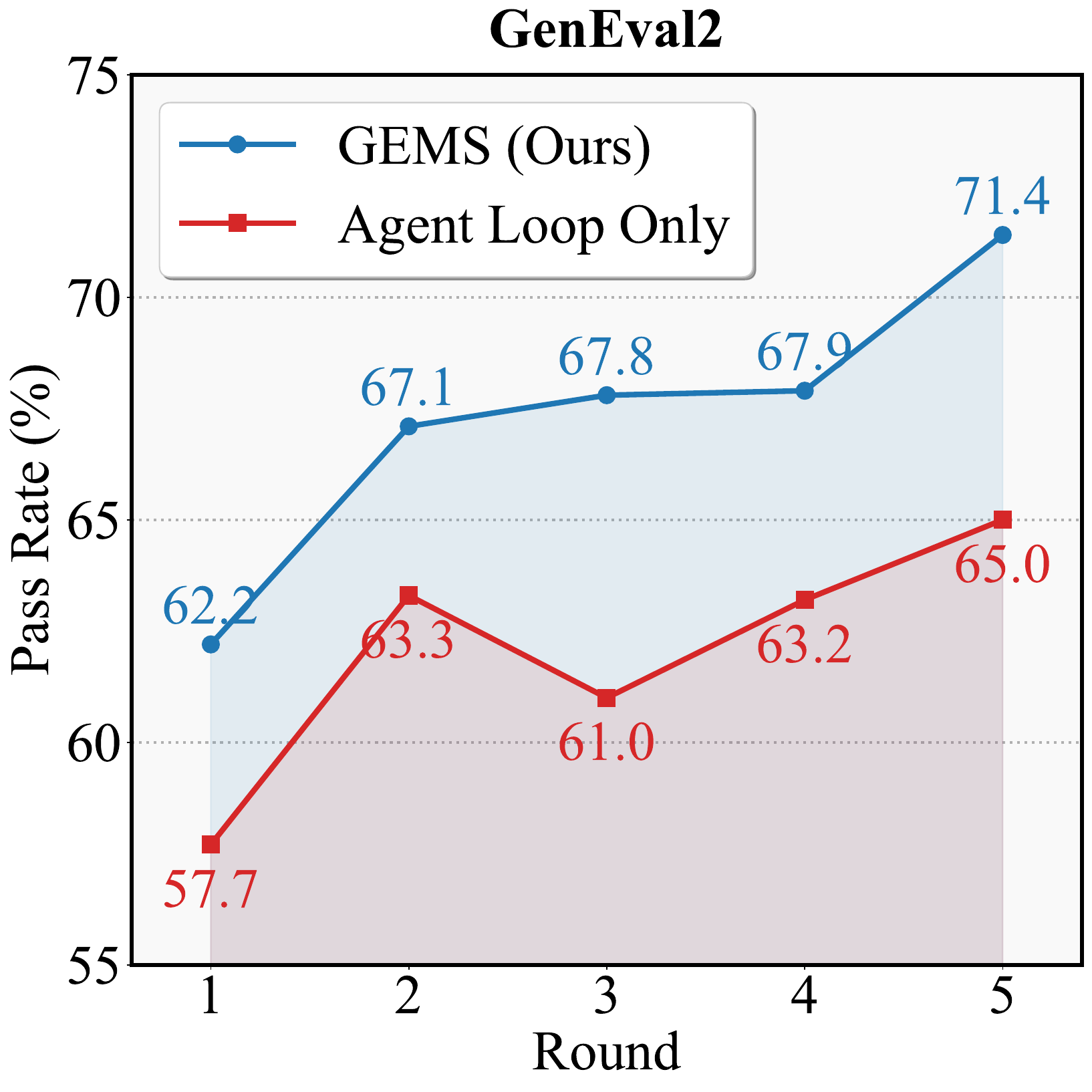}
        \label{fig:avg_pass_a}
    \end{subfigure}
    \hfill
    \begin{subfigure}{0.49\linewidth}
        \centering
        \includegraphics[width=\linewidth]{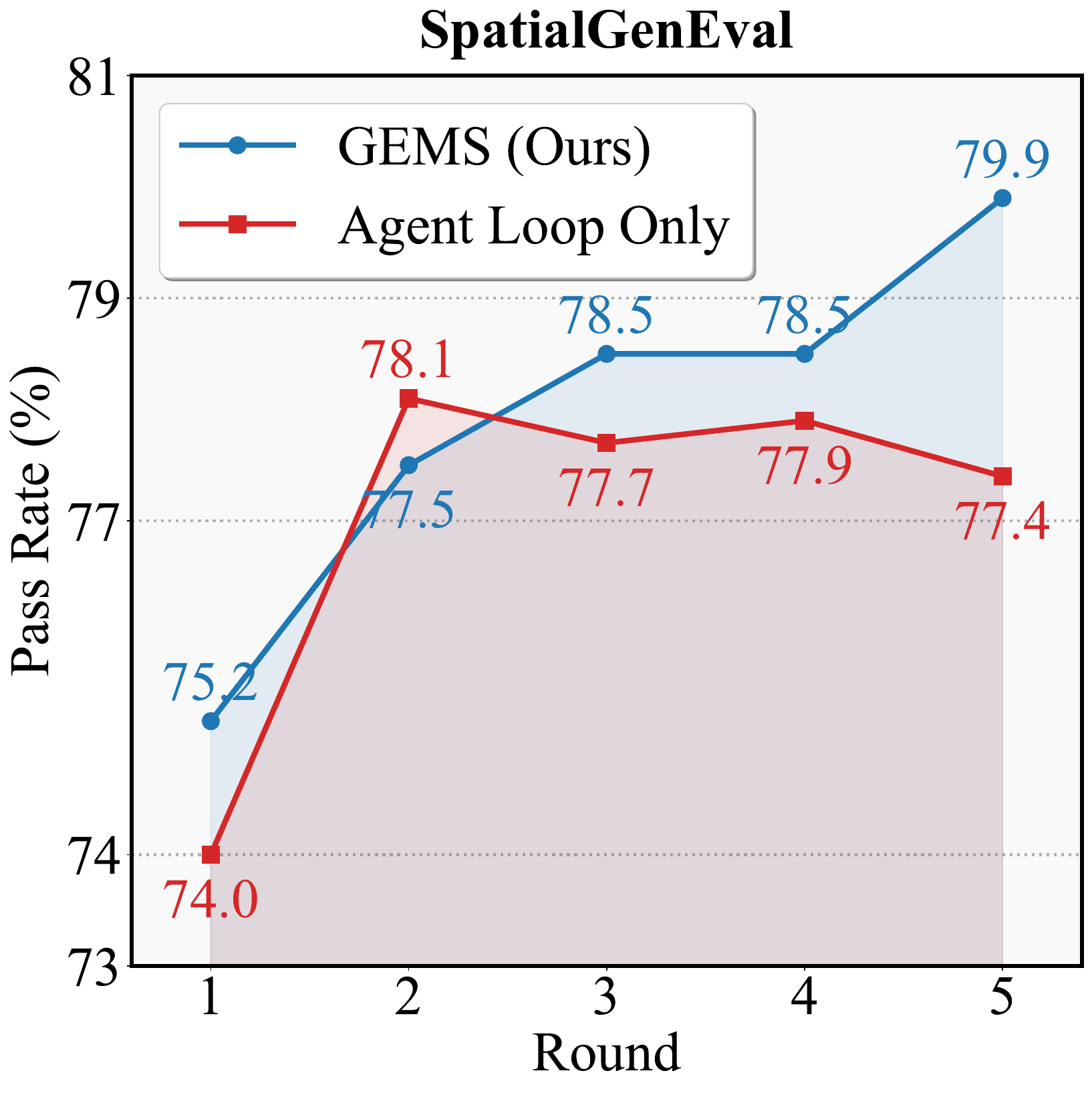}
        \label{fig:avg_pass_b}
    \end{subfigure}
    \vspace{-25pt}
        \caption{Average passed criteria over iterations.}
    \label{fig:avg_pass}
    \vspace{-25pt} 
\end{wrapfigure}

\paragraph{Analysis of Agent Loop}
As shown in Figure~\ref{fig:ablation_results}(left), the Agent Loop itself provides a substantial performance boost. A primary factor is the inherent stochasticity of the image generation process; within an iterative framework, as long as a single iteration produces a valid output, the Verifier can identify it as a success. In this sense, the loop partially functions like a Random Search~\cite{ma2025inference} strategy by providing multiple "shots" at the target. 

However, GEMS goes beyond mere repetition. To demonstrate that the prompt quality actually improves over time, we analyzed the average number of passed criteria across iterations on the most challenging benchmarks: GenEval2 and SpatialGenEval. As illustrated in Figure~\ref{fig:avg_pass}, while a basic Agent Loop Only approach shows some initial gains, its performance tends to fluctuate (e.g., in SpatialGenEval). In contrast, GEMS starts from a higher initial baseline and exhibits a consistent upward trajectory in success rate. For instance, on GenEval2, it progressively climbs from 62.2\% to 71.4\%, widening the margin over the baseline as rounds progress. This trend indicates that the Refiner is not merely generating random variations, but is actively performing directed optimization based on feedback. This ensures that GEMS fundamentally outperforms naive iterative methods.

\paragraph{Analysis of Agent Memory}
We further investigate the impact of different Agent Memory configurations, as illustrated in Figure~\ref{fig:ablation_results}(right). Initially, incorporating only historical prompts and their corresponding feedback leads to a 3.4 point improvement. Further including the generated images into the memory pool contributes an additional 3.1 points, suggesting that richer multimodal context provides more robust guidance for the refinement process. 

However, we find that more information is not always beneficial. When we attempt to include the full thought (CoT used to generate the corresponding prompt) into the memory, it results in negligible performance gains. We attribute this to the significant redundancy and informational noise present in raw reasoning logs~\cite{qu2025survey,sui2025stopoverthinkingsurveyefficient}, which can distract the Refiner or lead to token overhead. To address this, we utilize the Compressor to distill these raw thoughts into condensed "Experiences". This strategy yields a notable 2.5 point increase, confirming that concise, strategic insights are far more effective for long-context agentic reasoning than unprocessed internal reflections.

\begin{figure}[t] 
    \centering
    \begin{minipage}{0.59\textwidth}
        \centering
        \includegraphics[width=\linewidth]{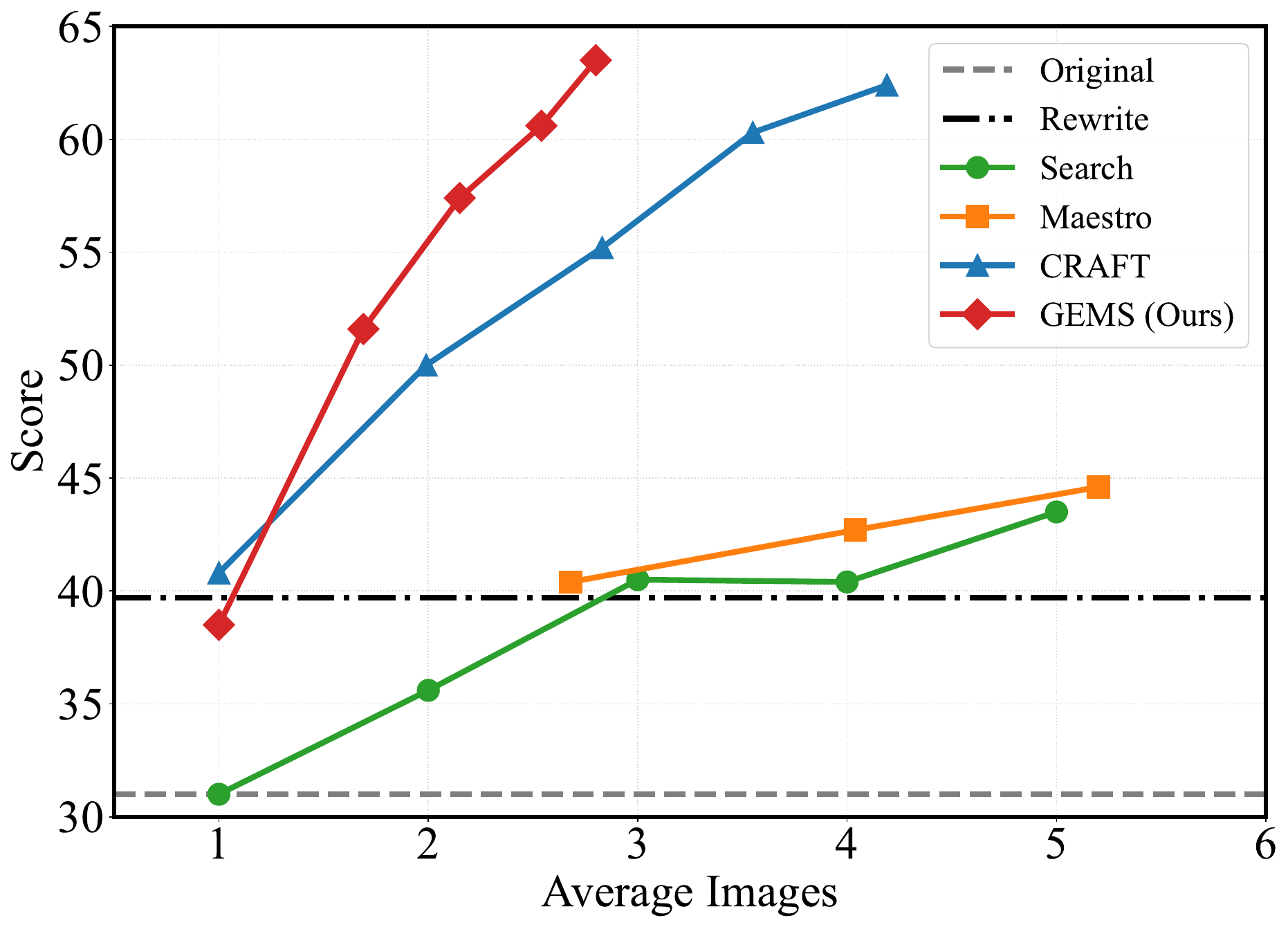}
        \caption{Trade-off between Efficiency and Performance. Comparison of inference-time scaling methods on GenEval2 using Z-Image-Turbo. GEMS (red line) achieves superior performance with fewer average images generated.}
        \label{fig:trade_off}
    \end{minipage}
    \hfill 
    \begin{minipage}{0.38\textwidth}
        \centering
        \includegraphics[width=\linewidth]{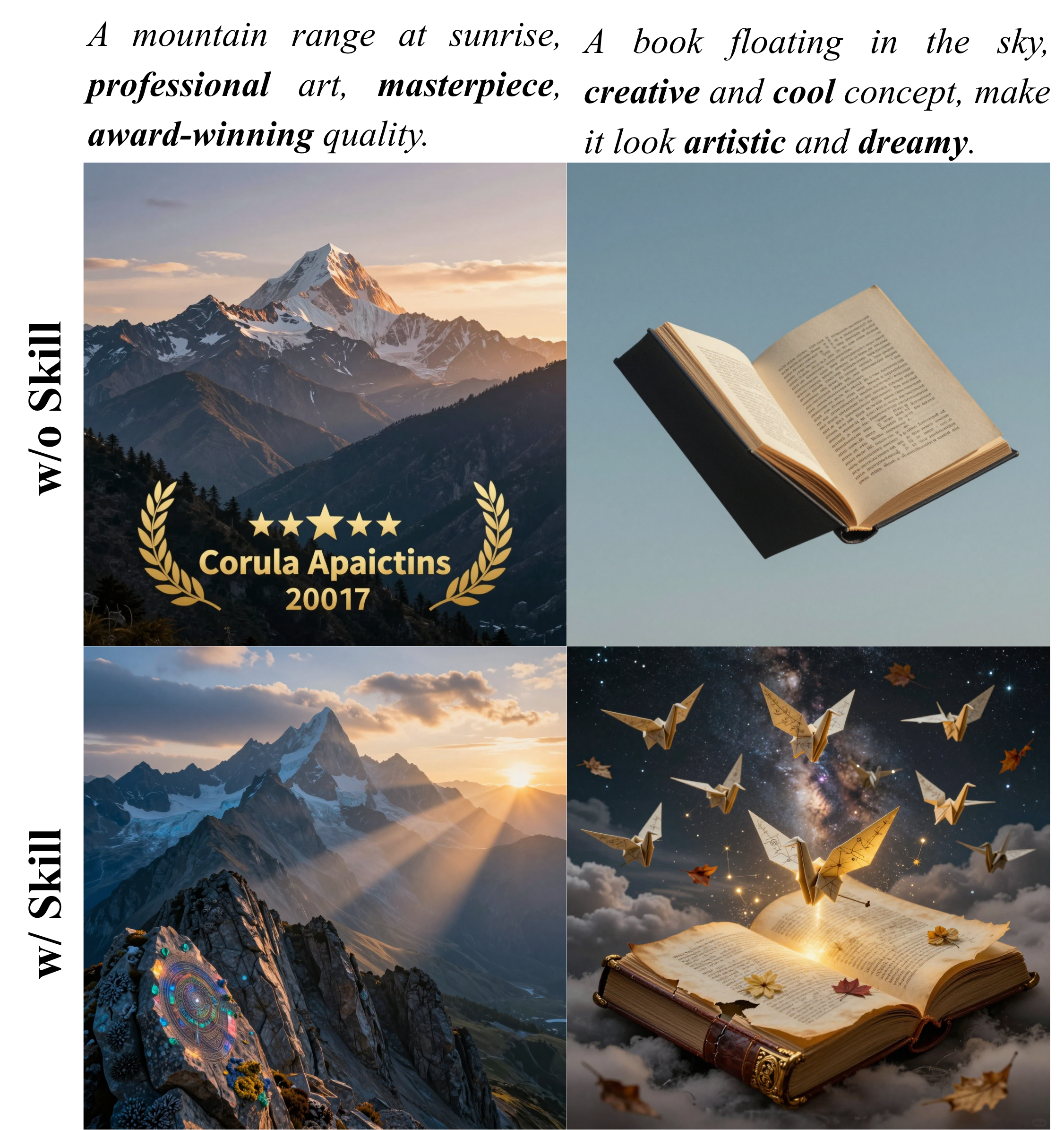}
        \caption{Qualitative Comparison of Agent Skills. GEMS autonomously triggers Aesthetic Drawing (left) and Creative Drawing (right) to enhance artistic expression.}
        \label{fig:skill_demo}
    \end{minipage}
    \vspace{-0pt} 
\end{figure}

\begin{figure}[t]
\centering
\includegraphics[width=\textwidth]{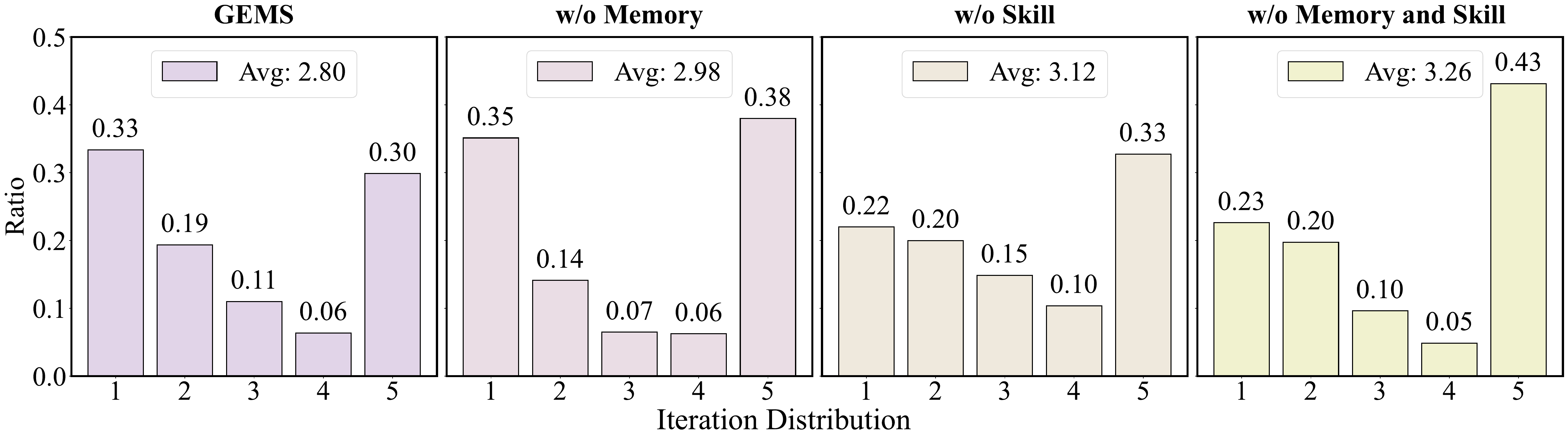}
\caption{Iteration distribution across GEMS ablation variants. By incorporating Memory and Skill, the system shifts the distribution toward earlier rounds and reducing the average number of iterations.}
\label{fig:distribute}
    \vspace{-10pt}
\end{figure}

\paragraph{Trade-off between Efficiency and Performance.} We also investigate the scaling behavior of various inference-time methods, focusing on the trade-off between computational efficiency and generative quality. As illustrated in Figure~\ref{fig:trade_off}, we evaluate these methods using Z-Image-Turbo on GenEval2, tracking the average number of generated images against the resulting scores. Due to the early stopping mechanism, GEMS delivers superior performance while maintaining significantly lower overhead. For instance, at an average of approximately three images per task, GEMS substantially outperforms other inference-time scaling methods. Further ablation in Figure~\ref{fig:distribute} shows that Agent Memory and Agent Skill enhance generation quality, therefore shifting the distribution toward earlier termination and reducing average iterations from 3.26 to 2.80.

\begin{figure}[t]
\centering
\includegraphics[width=\textwidth]{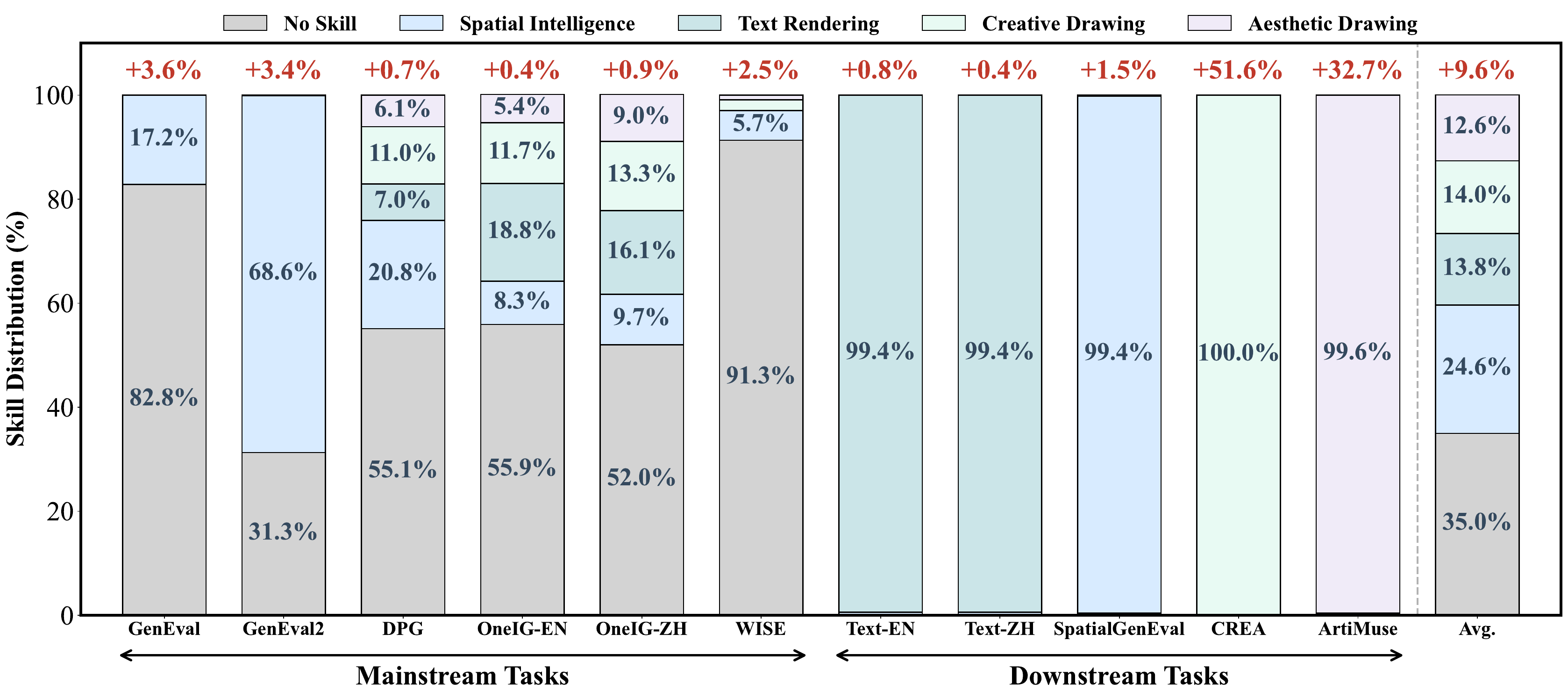}
\caption{Skill distribution across tasks and relative performance improvement. Each stacked bar illustrates the percentage composition of skills triggered within the respective benchmarks. The red numerical labels (e.g., +3.6\%) above the bars denote the relative performance gain achieved compared to the version without Agent Skill. }
\label{fig:skill_ab}
    \vspace{-10pt}
\end{figure}

\paragraph{Analysis of Agent Skill.} To further analyze the activation frequency of various skills and their impact on performance across different tasks, we provide a statistical breakdown of skill distribution and conduct an ablation study by specifically isolating the Agent Skill module, as illustrated in Figure~\ref{fig:skill_ab}. As observed, the system successfully invokes relevant domain-specific skills for downstream tasks; for instance, the {Spatial Intelligence} skill is predominantly triggered for SpatialGenEval, while the {Creative Drawing} skill is activated for CREA. We also illustrate the specific impact of the Creative Drawing and Aesthetic Drawing skills on the generated results, as depicted in Figure~\ref{fig:skill_demo}. By comparing outputs with and without these specialized skills, we observe that GEMS autonomously triggers specific Skills tailored to the user prompt, significantly improving overall visual appeal and composition quality. These results suggest that incorporating specific skills can effectively refine the visual quality of the generated content in downstream application.

Furthermore, the results demonstrate that skills are also selectively triggered within mainstream benchmarks. For example, general tasks involving spatial reasoning frequently activate the {Spatial Intelligence} skill. Specifically, breakdown results of GenEval (Detailed in Appendix~\ref{sec:result}, Table~\ref{tab:geneval_detailed}) reveal that the ``Position'' category exhibited the most pronounced improvements (+0.34), when using Z-Image-Turbo. Overall, these results demonstrate that Agent Skill improves performance on mainstream tasks by providing targeted enhancements in specific generative dimensions.

\section{Conclusion}
In this paper, we present GEMS, an agent-native multimodal generation framework that reframes text-to-image generation as an iterative optimization problem. By unifying state-of-the-art agentic practices into a framework with iterative refinement, persistent trajectory-level memory, and domain-specific skills, our approach enables progressive alignment between prompts and visual outputs, particularly for complex instructions and specialized downstream tasks.
Empirical results demonstrate that when integrated with Z-Image-Turbo, GEMS achieves significant performance gains of 14.22 across five mainstream benchmarks and 14.03 across four specialized downstream tasks. Notably, our framework surpasses the state-of-the-art Nano Banana 2 on GenEval2 with the 6B Z-Image-Turbo, demonstrating that agentic systems can effectively unlock the potential of smaller foundational models. We hope that GEMS will inspire subsequent explorations into agentic multimodal generation.

\bibliographystyle{abbrvnat}
\bibliography{main}

@article{seedream4,
  title={Seedream 4.0: Toward next-generation multimodal image generation},
  author={Seedream, Team and Chen, Yunpeng and Gao, Yu and Gong, Lixue and Guo, Meng and Guo, Qiushan and Guo, Zhiyao and Hou, Xiaoxia and Huang, Weilin and Huang, Yixuan and others},
  journal={arXiv preprint arXiv:2509.20427},
  year={2025}
}

@article{bagel,
  title={Emerging properties in unified multimodal pretraining},
  author={Deng, Chaorui and Zhu, Deyao and Li, Kunchang and Gou, Chenhui and Li, Feng and Wang, Zeyu and Zhong, Shu and Yu, Weihao and Nie, Xiaonan and Song, Ziang and others},
  journal={arXiv preprint arXiv:2505.14683},
  year={2025}
}

@article{z-image,
  title={Z-image: An efficient image generation foundation model with single-stream diffusion transformer},
  author={Cai, Huanqia and Cao, Sihan and Du, Ruoyi and Gao, Peng and Hoi, Steven and Hou, Zhaohui and Huang, Shijie and Jiang, Dengyang and Jin, Xin and Li, Liangchen and others},
  journal={arXiv preprint arXiv:2511.22699},
  year={2025}
}

@article{jiang2026xskill,
  title={XSkill: Continual Learning from Experience and Skills in Multimodal Agents},
  author={Jiang, Guanyu and Su, Zhaochen and Qu, Xiaoye and others},
  journal={arXiv preprint arXiv:2603.12056},
  year={2026}
}

@article{chen2025flash,
  title={Flash-DMD: Towards High-Fidelity Few-Step Image Generation with Efficient Distillation and Joint Reinforcement Learning},
  author={Chen, Guanjie and Huang, Shirui and Liu, Kai and Zhu, Jianchen and Qu, Xiaoye and Chen, Peng and Cheng, Yu and Sun, Yifu},
  journal={arXiv preprint arXiv:2511.20549},
  year={2025}
}

@article{qwen-image,
  title={Qwen-image technical report},
  author={Wu, Chenfei and Li, Jiahao and Zhou, Jingren and Lin, Junyang and Gao, Kaiyuan and Yan, Kun and Yin, Sheng-ming and Bai, Shuai and Xu, Xiao and Chen, Yilei and others},
  journal={arXiv preprint arXiv:2508.02324},
  year={2025}
}

@article{promptist,
  title={Optimizing prompts for text-to-image generation},
  author={Hao, Yaru and Chi, Zewen and Dong, Li and Wei, Furu},
  journal={Advances in Neural Information Processing Systems},
  volume={36},
  pages={66923--66939},
  year={2023}
}

@article{ma2025inference,
  title={Inference-time scaling for diffusion models beyond scaling denoising steps},
  author={Ma, Nanye and Tong, Shangyuan and Jia, Haolin and Hu, Hexiang and Su, Yu-Chuan and Zhang, Mingda and Yang, Xuan and Li, Yandong and Jaakkola, Tommi and Jia, Xuhui and others},
  journal={arXiv preprint arXiv:2501.09732},
  year={2025}
}

@article{maestro,
  title={Maestro: Self-improving text-to-image generation via agent orchestration},
  author={Wan, Xingchen and Zhou, Han and Sun, Ruoxi and Nakhost, Hootan and Jiang, Ke and Sinha, Rajarishi and Ar{\i}k, Sercan {\"O}},
  journal={arXiv preprint arXiv:2509.10704},
  year={2025}
}

@article{craft,
  title={CRAFT: Continuous Reasoning and Agentic Feedback Tuning for Multimodal Text-to-Image Generation},
  author={Kovalev, V and Kuvshinov, A and Buzovkin, A and Pokidov, D and Timonin, D},
  journal={arXiv preprint arXiv:2512.20362},
  year={2025}
}

@article{kimi2.5,
  title={Kimi K2. 5: Visual Agentic Intelligence},
  author={Team, Kimi and Bai, Tongtong and Bai, Yifan and Bao, Yiping and Cai, SH and Cao, Yuan and Charles, Y and Che, HS and Chen, Cheng and Chen, Guanduo and others},
  journal={arXiv preprint arXiv:2602.02276},
  year={2026}
}

@article{geneval,
  title={Geneval: An object-focused framework for evaluating text-to-image alignment},
  author={Ghosh, Dhruba and Hajishirzi, Hannaneh and Schmidt, Ludwig},
  journal={Advances in Neural Information Processing Systems},
  volume={36},
  pages={52132--52152},
  year={2023}
}

@article{geneval2,
  title={GenEval 2: Addressing Benchmark Drift in Text-to-Image Evaluation},
  author={Kamath, Amita and Chang, Kai-Wei and Krishna, Ranjay and Zettlemoyer, Luke and Hu, Yushi and Ghazvininejad, Marjan},
  journal={arXiv preprint arXiv:2512.16853},
  year={2025}
}

@article{dpg,
  title={Ella: Equip diffusion models with llm for enhanced semantic alignment},
  author={Hu, Xiwei and Wang, Rui and Fang, Yixiao and Fu, Bin and Cheng, Pei and Yu, Gang},
  journal={arXiv preprint arXiv:2403.05135},
  year={2024}
}

@article{wise,
  title={Wise: A world knowledge-informed semantic evaluation for text-to-image generation},
  author={Niu, Yuwei and Ning, Munan and Zheng, Mengren and Jin, Weiyang and Lin, Bin and Jin, Peng and Liao, Jiaqi and Feng, Chaoran and Ning, Kunpeng and Zhu, Bin and others},
  journal={arXiv preprint arXiv:2503.07265},
  year={2025}
}

@article{oneig,
  title={Oneig-bench: Omni-dimensional nuanced evaluation for image generation},
  author={Chang, Jingjing and Fang, Yixiao and Xing, Peng and Wu, Shuhan and Cheng, Wei and Wang, Rui and Zeng, Xianfang and Yu, Gang and Chen, Hai-Bao},
  journal={arXiv preprint arXiv:2506.07977},
  year={2025}
}

@article{longtextbench,
  title={X-omni: Reinforcement learning makes discrete autoregressive image generative models great again},
  author={Geng, Zigang and Wang, Yibing and Ma, Yeyao and Li, Chen and Rao, Yongming and Gu, Shuyang and Zhong, Zhao and Lu, Qinglin and Hu, Han and Zhang, Xiaosong and others},
  journal={arXiv preprint arXiv:2507.22058},
  year={2025}
}

@article{spatialgeneval,
  title={Everything in Its Place: Benchmarking Spatial Intelligence of Text-to-Image Models},
  author={Wang, Zengbin and Hu, Xuecai and Wang, Yong and Xiong, Feng and Zhang, Man and Chu, Xiangxiang},
  journal={arXiv preprint arXiv:2601.20354},
  year={2026}
}

@article{crea,
  title={CREA: A Collaborative Multi-Agent Framework for Creative Image Editing and Generation},
  author={Venkatesh, Kavana and Dunlop, Connor and Yanardag, Pinar},
  journal={arXiv preprint arXiv:2504.05306},
  year={2025}
}

@article{fang2025got,
  title={Got: Unleashing reasoning capability of multimodal large language model for visual generation and editing},
  author={Fang, Rongyao and Duan, Chengqi and Wang, Kun and Huang, Linjiang and Li, Hao and Yan, Shilin and Tian, Hao and Zeng, Xingyu and Zhao, Rui and Dai, Jifeng and others},
  journal={arXiv preprint arXiv:2503.10639},
  year={2025}
}

@article{jiang2025t2i,
  title={T2i-r1: Reinforcing image generation with collaborative semantic-level and token-level cot},
  author={Jiang, Dongzhi and Guo, Ziyu and Zhang, Renrui and Zong, Zhuofan and Li, Hao and Zhuo, Le and Yan, Shilin and Heng, Pheng-Ann and Li, Hongsheng},
  journal={arXiv preprint arXiv:2505.00703},
  year={2025}
}

@inproceedings{liao2025imagegen,
  title={Imagegen-cot: Enhancing text-to-image in-context learning with chain-of-thought reasoning},
  author={Liao, Jiaqi and Yang, Zhengyuan and Li, Linjie and Li, Dianqi and Lin, Kevin and Cheng, Yu and Wang, Lijuan},
  booktitle={Proceedings of the IEEE/CVF International Conference on Computer Vision},
  pages={17214--17223},
  year={2025}
}

@article{jiao2025thinkgen,
  title={ThinkGen: Generalized Thinking for Visual Generation},
  author={Jiao, Siyu and Lin, Yiheng and Zhong, Yujie and She, Qi and Zhou, Wei and Lan, Xiaohan and Huang, Zilong and Yu, Fei and Yu, Yingchen and Zhao, Yunqing and others},
  journal={arXiv preprint arXiv:2512.23568},
  year={2025}
}

@article{kou2026think,
  title={Think-Then-Generate: Reasoning-Aware Text-to-Image Diffusion with LLM Encoders},
  author={Kou, Siqi and Jin, Jiachun and Zhou, Zetong and Ma, Ye and Wang, Yugang and Chen, Quan and Jiang, Peng and Yang, Xiao and Zhu, Jun and Yu, Kai and others},
  journal={arXiv preprint arXiv:2601.10332},
  year={2026}
}

@article{wang2025imagent,
  title={ImAgent: A Unified Multimodal Agent Framework for Test-Time Scalable Image Generation},
  author={Wang, Kaishen and Chen, Ruibo and Zheng, Tong and Huang, Heng},
  journal={arXiv preprint arXiv:2511.11483},
  year={2025}
}

@article{jiang2026genagent,
  title={GenAgent: Scaling Text-to-Image Generation via Agentic Multimodal Reasoning},
  author={Jiang, Kaixun and Wang, Yuzheng and Zhou, Junjie and Li, Pandeng and Liu, Zhihang and Xie, Chen-Wei and Chen, Zhaoyu and Zheng, Yun and Zhang, Wenqiang},
  journal={arXiv preprint arXiv:2601.18543},
  year={2026}
}

@article{mondal2025countloop,
  title={CountLoop: Training-Free High-Instance Image Generation via Iterative Agent Guidance},
  author={Mondal, Anindya and Banerjee, Ayan and Nag, Sauradip and Llad{\~A}{\`g}s, Josep and Zhu, Xiatian and Dutta, Anjan},
  journal={arXiv preprint arXiv:2508.16644},
  year={2025}
}

@inproceedings{li2025reflect,
  title={Reflect-dit: Inference-time scaling for text-to-image diffusion transformers via in-context reflection},
  author={Li, Shufan and Kallidromitis, Konstantinos and Gokul, Akash and Koneru, Arsh and Kato, Yusuke and Kozuka, Kazuki and Grover, Aditya},
  booktitle={Proceedings of the IEEE/CVF International Conference on Computer Vision},
  pages={15657--15668},
  year={2025}
}

@inproceedings{zhuo2025reflection,
  title={From reflection to perfection: Scaling inference-time optimization for text-to-image diffusion models via reflection tuning},
  author={Zhuo, Le and Zhao, Liangbing and Paul, Sayak and Liao, Yue and Zhang, Renrui and Xin, Yi and Gao, Peng and Elhoseiny, Mohamed and Li, Hongsheng},
  booktitle={Proceedings of the IEEE/CVF International Conference on Computer Vision},
  pages={15329--15339},
  year={2025}
}

@article{li2025editthinker,
  title={Editthinker: Unlocking iterative reasoning for any image editor},
  author={Li, Hongyu and Zhang, Manyuan and Zheng, Dian and Guo, Ziyu and Jia, Yimeng and Feng, Kaituo and Yu, Hao and Liu, Yexin and Feng, Yan and Pei, Peng and others},
  journal={arXiv preprint arXiv:2512.05965},
  year={2025}
}

@article{guo2025thinking,
  title={Thinking-while-generating: Interleaving textual reasoning throughout visual generation},
  author={Guo, Ziyu and Zhang, Renrui and Li, Hongyu and Zhang, Manyuan and Chen, Xinyan and Wang, Sifan and Feng, Yan and Pei, Peng and Heng, Pheng-Ann},
  journal={arXiv preprint arXiv:2511.16671},
  year={2025}
}

@inproceedings{li2025mccd,
  title={Mccd: Multi-agent collaboration-based compositional diffusion for complex text-to-image generation},
  author={Li, Mingcheng and Hou, Xiaolu and Liu, Ziyang and Yang, Dingkang and Qian, Ziyun and Chen, Jiawei and Wei, Jinjie and Jiang, Yue and Xu, Qingyao and Zhang, Lihua},
  booktitle={Proceedings of the Computer Vision and Pattern Recognition Conference},
  pages={13263--13272},
  year={2025}
}

@inproceedings{park2026guiding,
  title={Guiding What Not to Generate: Automated Negative Prompting for Text-Image Alignment},
  author={Park, Sangha and Kim, Eunji and Oh, Yeongtak and Choi, Jooyoung and Yoon, Sungroh},
  booktitle={Proceedings of the IEEE/CVF Winter Conference on Applications of Computer Vision},
  pages={6664--6675},
  year={2026}
}

@article{zhu2026paperbanana,
  title={PaperBanana: Automating Academic Illustration for AI Scientists},
  author={Zhu, Dawei and Meng, Rui and Song, Yale and Wei, Xiyu and Li, Sujian and Pfister, Tomas and Yoon, Jinsung},
  journal={arXiv preprint arXiv:2601.23265},
  year={2026}
}

@article{ramesh2022hierarchical,
  title={Hierarchical text-conditional image generation with clip latents},
  author={Ramesh, Aditya and Dhariwal, Prafulla and Nichol, Alex and Chu, Casey and Chen, Mark},
  journal={arXiv preprint arXiv:2204.06125},
  volume={1},
  number={2},
  pages={3},
  year={2022}
}

@article{saharia2022photorealistic,
  title={Photorealistic text-to-image diffusion models with deep language understanding},
  author={Saharia, Chitwan and Chan, William and Saxena, Saurabh and Li, Lala and Whang, Jay and Denton, Emily L and Ghasemipour, Kamyar and Gontijo Lopes, Raphael and Karagol Ayan, Burcu and Salimans, Tim and others},
  journal={Advances in neural information processing systems},
  volume={35},
  pages={36479--36494},
  year={2022}
}

@article{brooks2024video,
  title={Video generation models as world simulators},
  author={Brooks, Tim and Peebles, Bill and Holmes, Connor and DePue, Will and Guo, Yufei and Jing, Li and Schnurr, David and Taylor, Joe and Luhman, Troy and Luhman, Eric and others},
  journal={OpenAI Blog},
  volume={1},
  number={8},
  pages={1},
  year={2024}
}

@article{he2025diffthinker,
  title={Diffthinker: Towards generative multimodal reasoning with diffusion models},
  author={He, Zefeng and Qu, Xiaoye and Li, Yafu and Zhu, Tong and Huang, Siyuan and Cheng, Yu},
  journal={arXiv preprint arXiv:2512.24165},
  year={2025}
}

@inproceedings{wei2022chain,
  title={Chain-of-Thought Prompting Elicits Reasoning in Large Language Models},
  author={Wei, Jason and Wang, Xuezhi and Schuurmans, Dale and Bosma, Maarten and Fei-Fei, Li and Chi, Ed and Le, Quoc V and Zhou, Denny},
  booktitle={Advances in Neural Information Processing Systems (NeurIPS)},
  volume={35},
  year={2022}
}

@inproceedings{yao2022react,
  title={React: Synergizing reasoning and acting in language models},
  author={Yao, Shunyu and Zhao, Jeffrey and Yu, Dian and Du, Nan and Shafran, Izhak and Narasimhan, Karthik R and Cao, Yuan},
  booktitle={The eleventh international conference on learning representations},
  year={2022}
}

@article{shinn2023reflexion,
  title={Reflexion: Language agents with verbal reinforcement learning},
  author={Shinn, Noah and Cassano, Federico and Gopinath, Ashwin and Narasimhan, Karthik and Yao, Shunyu},
  journal={Advances in neural information processing systems},
  volume={36},
  pages={8634--8652},
  year={2023}
}

@inproceedings{wang2023plan,
  title={Plan-and-solve prompting: Improving zero-shot chain-of-thought reasoning by large language models},
  author={Wang, Lei and Xu, Wanyu and Lan, Yihuai and Hu, Zhiqiang and Lan, Yunshi and Lee, Roy Ka-Wei and Lim, Ee-Peng},
  booktitle={Proceedings of the 61st annual meeting of the association for computational linguistics (volume 1: long papers)},
  pages={2609--2634},
  year={2023}
}

@article{madaan2023self,
  title={Self-refine: Iterative refinement with self-feedback},
  author={Madaan, Aman and Tandon, Niket and Gupta, Prakhar and Hallinan, Skyler and Gao, Luyu and Wiegreffe, Sarah and Alon, Uri and Dziri, Nouha and Prabhumoye, Shrimai and Yang, Yiming and others},
  journal={Advances in neural information processing systems},
  volume={36},
  pages={46534--46594},
  year={2023}
}

@article{schick2023toolformer,
  title={Toolformer: Language models can teach themselves to use tools},
  author={Schick, Timo and Dwivedi-Yu, Jane and Dess{\`\i}, Roberto and Raileanu, Roberta and Lomeli, Maria and Hambro, Eric and Zettlemoyer, Luke and Cancedda, Nicola and Scialom, Thomas},
  journal={Advances in neural information processing systems},
  volume={36},
  pages={68539--68551},
  year={2023}
}

@article{he2025framethinker,
  title={Framethinker: Learning to think with long videos via multi-turn frame spotlighting},
  author={He, Zefeng and Qu, Xiaoye and Li, Yafu and Huang, Siyuan and Liu, Daizong and Cheng, Yu},
  journal={arXiv preprint arXiv:2509.24304},
  year={2025}
}

@article{li2023camel,
  title={Camel: Communicative agents for" mind" exploration of large language model society},
  author={Li, Guohao and Hammoud, Hasan and Itani, Hani and Khizbullin, Dmitrii and Ghanem, Bernard},
  journal={Advances in neural information processing systems},
  volume={36},
  pages={51991--52008},
  year={2023}
}

@inproceedings{hong2023metagpt,
  title={MetaGPT: Meta programming for a multi-agent collaborative framework},
  author={Hong, Sirui and Zhuge, Mingchen and Chen, Jonathan and Zheng, Xiawu and Cheng, Yuheng and Wang, Jinlin and Zhang, Ceyao and Wang, Zili and Yau, Steven Ka Shing and Lin, Zijuan and others},
  booktitle={The twelfth international conference on learning representations},
  year={2023}
}

@inproceedings{wu2024autogen,
  title={Autogen: Enabling next-gen LLM applications via multi-agent conversations},
  author={Wu, Qingyun and Bansal, Gagan and Zhang, Jieyu and Wu, Yiran and Li, Beibin and Zhu, Erkang and Jiang, Li and Zhang, Xiaoyun and Zhang, Shaokun and Liu, Jiale and others},
  booktitle={First conference on language modeling},
  year={2024}
}

@inproceedings{chen2023agentverse,
  title={Agentverse: Facilitating multi-agent collaboration and exploring emergent behaviors},
  author={Chen, Weize and Su, Yusheng and Zuo, Jingwei and Yang, Cheng and Yuan, Chenfei and Chan, Chi-Min and Yu, Heyang and Lu, Yaxi and Hung, Yi-Hsin and Qian, Chen and others},
  booktitle={The Twelfth International Conference on Learning Representations},
  year={2023}
}

@article{packer2023memgpt,
  title={MemGPT: towards LLMs as operating systems.},
  author={Packer, Charles and Fang, Vivian and Patil, Shishir\_G and Lin, Kevin and Wooders, Sarah and Gonzalez, Joseph\_E},
  journal={arXiv preprint arXiv:2310.08560},
  year={2023},
}

@article{chhikara2025mem0,
  title={Mem0: Building production-ready ai agents with scalable long-term memory},
  author={Chhikara, Prateek and Khant, Dev and Aryan, Saket and Singh, Taranjeet and Yadav, Deshraj},
  journal={arXiv preprint arXiv:2504.19413},
  year={2025}
}

@article{xu2025mem,
  title={A-mem: Agentic memory for llm agents},
  author={Xu, Wujiang and Liang, Zujie and Mei, Kai and Gao, Hang and Tan, Juntao and Zhang, Yongfeng},
  journal={arXiv preprint arXiv:2502.12110},
  year={2025}
}

@article{long2025seeing,
  title={Seeing, listening, remembering, and reasoning: A multimodal agent with long-term memory},
  author={Long, Lin and He, Yichen and Ye, Wentao and Pan, Yiyuan and Lin, Yuan and Li, Hang and Zhao, Junbo and Li, Wei},
  journal={arXiv preprint arXiv:2508.09736},
  year={2025}
}

@article{fu2026latentmem,
  title={LatentMem: Customizing Latent Memory for Multi-Agent Systems},
  author={Fu, Muxin and Zhang, Guibin and Xue, Xiangyuan and Li, Yafu and He, Zefeng and Huang, Siyuan and Qu, Xiaoye and Cheng, Yu and Yang, Yang},
  journal={arXiv preprint arXiv:2602.03036},
  year={2026}
}

@article{xia2026skillrl,
  title={SkillRL: Evolving Agents via Recursive Skill-Augmented Reinforcement Learning},
  author={Xia, Peng and Chen, Jianwen and Wang, Hanyang and Liu, Jiaqi and Zeng, Kaide and Wang, Yu and Han, Siwei and Zhou, Yiyang and Zhao, Xujiang and Chen, Haifeng and others},
  journal={arXiv preprint arXiv:2602.08234},
  year={2026}
}

@article{liang2026skillnet,
  title={SkillNet: Create, Evaluate, and Connect AI Skills},
  author={Liang, Yuan and Zhong, Ruobin and Xu, Haoming and Jiang, Chen and Zhong, Yi and Fang, Runnan and Gu, Jia-Chen and Deng, Shumin and Yao, Yunzhi and Wang, Mengru and others},
  journal={arXiv preprint arXiv:2603.04448},
  year={2026}
}

@article{li2026skillsbench,
  title={SkillsBench: Benchmarking how well agent skills work across diverse tasks},
  author={Li, Xiangyi and Chen, Wenbo and Liu, Yimin and Zheng, Shenghan and Chen, Xiaokun and He, Yifeng and Li, Yubo and You, Bingran and Shen, Haotian and Sun, Jiankai and others},
  journal={arXiv preprint arXiv:2602.12670},
  year={2026}
}

@article{qu2025survey,
  title={A survey of efficient reasoning for large reasoning models: Language, multimodality, and beyond},
  author={Qu, Xiaoye and Li, Yafu and Su, Zhaochen and Sun, Weigao and Yan, Jianhao and Liu, Dongrui and Cui, Ganqu and Liu, Daizong and Liang, Shuxian and He, Junxian and others},
  journal={arXiv preprint arXiv:2503.21614},
  year={2025}
}

@article{sui2025stopoverthinkingsurveyefficient,
      title={Stop Overthinking: A Survey on Efficient Reasoning for Large Language Models}, 
      author={Yang Sui and Yu-Neng Chuang and Guanchu Wang and Jiamu Zhang and Tianyi Zhang and Jiayi Yuan and Hongyi Liu and Andrew Wen and Shaochen Zhong and Hanjie Chen and Xia Hu},
      journal={arXiv preprint arXiv:2503.16419},
      year={2025},
}

@article{ho2020denoising,
  title={Denoising diffusion probabilistic models},
  author={Ho, Jonathan and Jain, Ajay and Abbeel, Pieter},
  journal={Advances in neural information processing systems},
  volume={33},
  pages={6840--6851},
  year={2020}
}

@article{song2020denoising,
  title={Denoising diffusion implicit models},
  author={Song, Jiaming and Meng, Chenlin and Ermon, Stefano},
  journal={arXiv preprint arXiv:2010.02502},
  year={2020}
}

@article{ho2022classifier,
  title={Classifier-free diffusion guidance},
  author={Ho, Jonathan and Salimans, Tim},
  journal={arXiv preprint arXiv:2207.12598},
  year={2022}
}

@article{lipman2022flow,
  title={Flow matching for generative modeling},
  author={Lipman, Yaron and Chen, Ricky TQ and Ben-Hamu, Heli and Nickel, Maximilian and Le, Matt},
  journal={arXiv preprint arXiv:2210.02747},
  year={2022}
}

@article{liu2022flow,
  title={Flow straight and fast: Learning to generate and transfer data with rectified flow},
  author={Liu, Xingchao and Gong, Chengyue and Liu, Qiang},
  journal={arXiv preprint arXiv:2209.03003},
  year={2022}
}

@article{albergo2022building,
  title={Building normalizing flows with stochastic interpolants},
  author={Albergo, Michael S and Vanden-Eijnden, Eric},
  journal={arXiv preprint arXiv:2209.15571},
  year={2022}
}

@inproceedings{rombach2022high,
  title={High-resolution image synthesis with latent diffusion models},
  author={Rombach, Robin and Blattmann, Andreas and Lorenz, Dominik and Esser, Patrick and Ommer, Bj{\"o}rn},
  booktitle={Proceedings of the IEEE/CVF conference on computer vision and pattern recognition},
  pages={10684--10695},
  year={2022}
}

@inproceedings{peebles2023scalable,
  title={Scalable diffusion models with transformers},
  author={Peebles, William and Xie, Saining},
  booktitle={Proceedings of the IEEE/CVF international conference on computer vision},
  pages={4195--4205},
  year={2023}
}

@article{mmdit,
  title={Scaling Rectified Flow Transformers for High-Resolution Image Synthesis},
  author={Esser, Patrick and Kulal, Sumith and Andreas, Andreas and Levi, Levi and Chertok, Michael and Gallo, Harry and Ganguli, Deep and Chou, Kyle and Kim, Sung and Crowson, Katherine and others},
  journal={arXiv preprint arXiv:2403.03206},
  year={2024}
}

@article{cao2025artimuse,
  title={Artimuse: Fine-grained image aesthetics assessment with joint scoring and expert-level understanding},
  author={Cao, Shuo and Ma, Nan and Li, Jiayang and Li, Xiaohui and Shao, Lihao and Zhu, Kaiwen and Zhou, Yu and Pu, Yuandong and Wu, Jiarui and Wang, Jiaquan and others},
  journal={arXiv preprint arXiv:2507.14533},
  year={2025}
}

@article{cao2025unipercept,
  title={UniPercept: Towards Unified Perceptual-Level Image Understanding across Aesthetics, Quality, Structure, and Texture},
  author={Cao, Shuo and Li, Jiayang and Li, Xiaohui and Pu, Yuandong and Zhu, Kaiwen and Gao, Yuanting and Luo, Siqi and Xin, Yi and Qin, Qi and Zhou, Yu and others},
  journal={arXiv preprint arXiv:2512.21675},
  year={2025}
}

@misc{glm-image,
  author  = {Z.ai},
  title   = {GLM-Image: Auto-regressive for Dense-knowledge and High-fidelity Image Generation},
  year    = {2026},
  url     = {https://z.ai/blog/glm-image},
  urldate = {2026-01-14}
}


\appendix

\clearpage

\section{Experiment Details}
\subsection{Prompt}
\begin{figure}[t]
\centering
\vspace{-10pt}
\includegraphics[width=\textwidth]{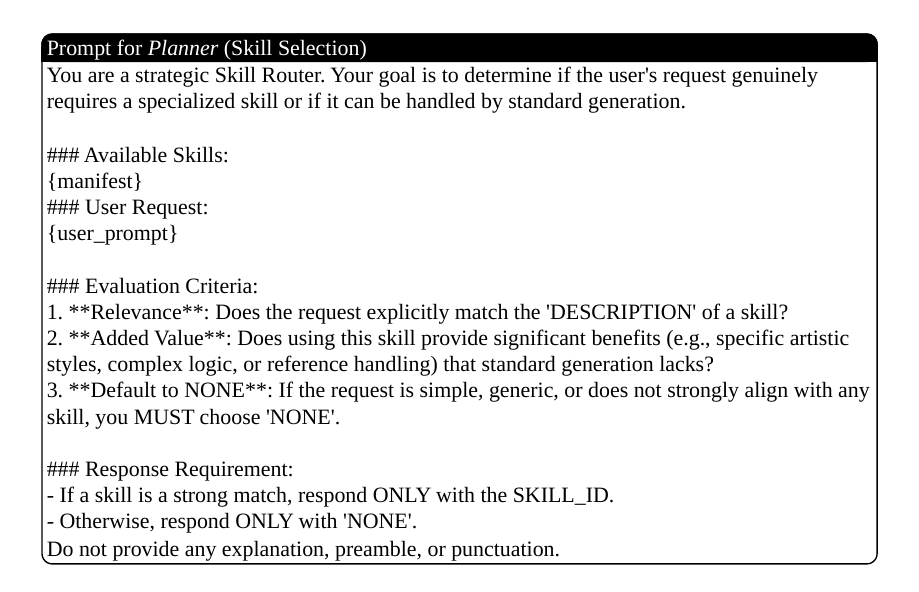}
\vspace{-20pt}
\caption{{Prompt for Planner (Skill Selection).}}
\label{fig:Planner1}
\end{figure}

\begin{figure}[t]
\centering
\vspace{-10pt}
\includegraphics[width=\textwidth]{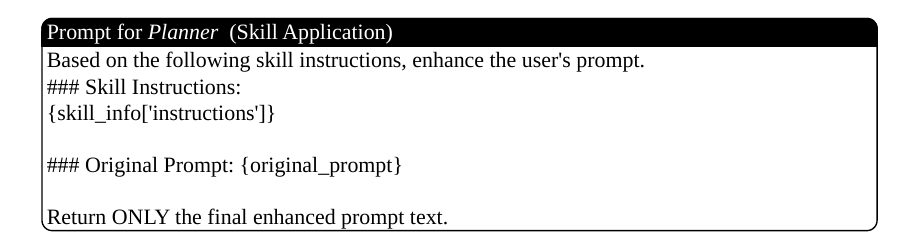}
\vspace{-20pt}
\caption{{Prompt for Planner (Skill Application).}}
\label{fig:Planner2}
\end{figure}

\begin{figure}[t]
\centering
\vspace{-10pt}
\includegraphics[width=\textwidth]{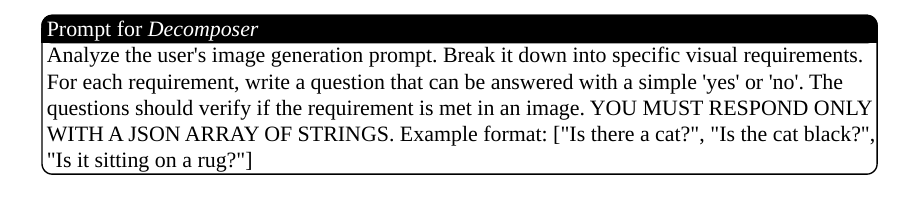}
\vspace{-20pt}
\caption{{Prompt for Decomposer.}}
\label{fig:Decomposer}
\end{figure}

\begin{figure}[t]
\centering
\vspace{-10pt}
\includegraphics[width=\textwidth]{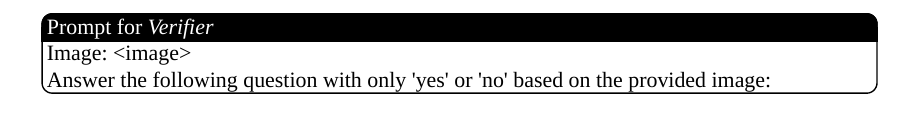}
\caption{{Prompt for Verifier.}}
\label{fig:Verifier}
\end{figure}

\begin{figure}[t]
\centering
\vspace{-10pt}
\includegraphics[width=\textwidth]{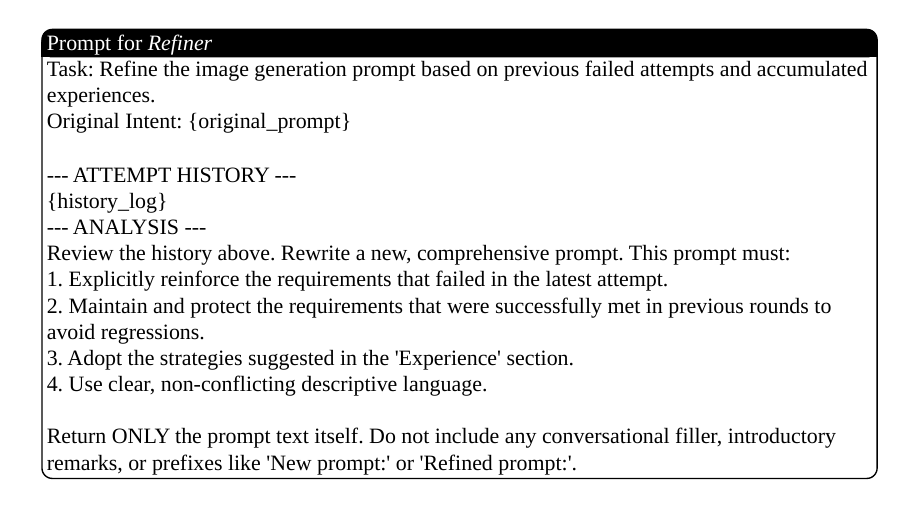}
\caption{{Prompt for Refiner.}}
\label{fig:Refiner}
\end{figure}

\begin{figure}[t]
\centering
\vspace{-10pt}
\includegraphics[width=\textwidth]{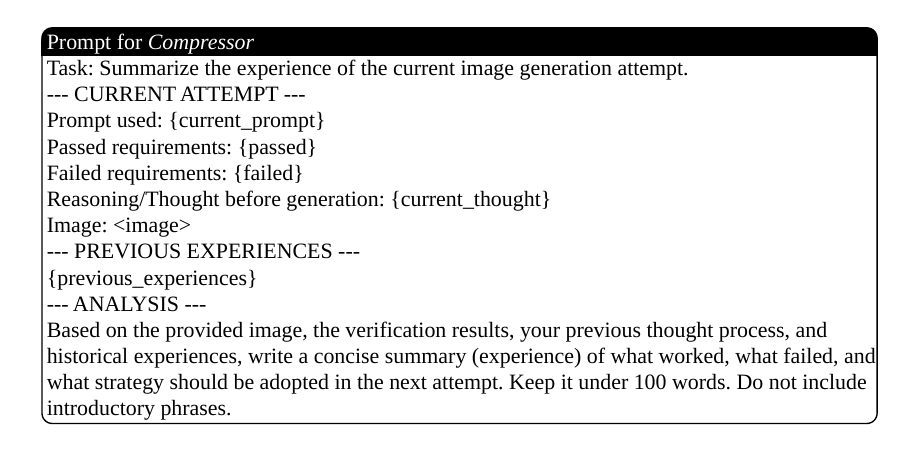}
\caption{{Prompt for Compressor.}}
\label{fig:Compressor}
\end{figure}

The detailed prompts for the LLM-based modules (Planner, Decomposer, Verifier, Refiner, and Compressor) are presented in Figures~\ref{fig:Planner1} through~\ref{fig:Compressor}. In contrast, the Skill Manager and Memory Manager are implemented as programmatic modules (e.g., Python class) rather than LLM agents.

\clearpage

\subsection{Evaluation Details}
\label{sec:eval}
\noindent 
\begin{minipage}[t]{0.48\textwidth} 
    \vspace{0pt} 
We conduct our evaluations following the official settings of Z-Image~\cite{z-image} and Qwen-Image~\cite{qwen-image}. For a comprehensive comparison, the configuration of the state-of-the-art Nano Banana 2 is also presented. Detailed specifications for these models are summarized in Table~\ref{tab:model_specs}.
\end{minipage}
\hfill 
\begin{minipage}[t]{0.48\textwidth} 
    \centering
    \captionof{table}{{Evaluation Details.}} 
    \label{tab:model_specs}
    \small
    \resizebox{\columnwidth}{!}{%
    \begin{tabular}{lcc}
\toprule
\textbf{Model} & \textbf{Resolution} & \textbf{DiT Forwards} \\ \midrule
Z-Image-Turbo & $1024 \times 1024$ & 8 \\
Qwen-Image-2512 & $1328 \times 1328$ & 50 \\ 
Nano Banana 2 & $1024 \times 1024$ & -- \\
\bottomrule
\end{tabular}
}
\end{minipage}

\subsection{Benchmark Details}
\label{sec:benchmark}
The four downstream tasks (LongText-Bench, SpatialGenEval, CREA, and ArtiMuse) primarily focus on evaluating the model's capabilities in text rendering, spatial intelligence, creative drawing, and aesthetic drawing, respectively.
For ArtiMuse test, we employ the ArtiMuse~\cite{cao2025artimuse} model as our dedicated scoring metric to evaluate the aesthetic quality of the generated images. Following the experimental protocol in Unipercept~\cite{cao2025unipercept}, we adopt prompts from the GenEval benchmark. To ensure the generative systems are evaluated at their aesthetic potential, we prepend the prefix ``Aesthetic Drawing: '' to each prompt, thereby steering the models toward generating results with higher visual appeal for ArtiMuse to assess. Regarding CREA~\cite{crea}, since it is not fully open-sourced, we perform our evaluation using half of its publicly available data. We employ Kimi as the judge. For all other benchmarks, we strictly adhere to the official scripts.

\subsection{Baseline Details}
\label{sec:baseline}

In this section, we provide details of the inference-time scaling baselines compared in our study.

\paragraph{Rewrite.}
The Rewrite baseline employs the MLLM to enhance initial user prompts following established best practices from the Google Cloud text-to-image prompt guide.

\paragraph{Promptist~\cite{promptist}.}
The Promptist baseline utilizes a fine-tuned language model to automatically rephrase user prompts. It generates optimized refinements via beam search, aiming to better align user intent with the capabilities of text-to-image models.

\paragraph{Random Search~\cite{ma2025inference}.} 
The Random Search baseline generates multiple images and selects the best based on scores assigned by a verifier. In our implementation, we utilize a MLLM as verifier.

\paragraph{Maestro~\cite{maestro}.}
Maestro is a self-improving text-to-image framework that utilizes iterative generation, producing two images in each round and featuring a pairwise comparison mechanism to evolve. Since the official code is not publicly available, we re-implemented the framework  as a baseline based on the original paper, specifically incorporating its signature pairwise comparison logic.

\paragraph{CRAFT~\cite{craft}.}
CRAFT is an iterative framework that refines prompts based on MLLM-based feedback. It evaluates images against generated visual questions and performs targeted prompt updates to address identified failures, without employing specialized memory management.

\section{Result Details}
\label{sec:result}
In this section, we provide a fine-grained performance breakdown for GenEval~\cite{geneval}, GenEval2~\cite{geneval2}, DPG-Bench~\cite{dpg}, OneIG~\cite{oneig}, WISE~\cite{wise}, SpatialGenEval~\cite{spatialgeneval}, and CREA~\cite{crea}. Since these benchmarks categorize prompts into diverse sub-dimensions, we report the scores for each individual subset adhering to the official scripts. These detailed results offer a more comprehensive comparison of how different inference-time scaling strategies impact specific generation capabilities.

\begin{table*}[htbp]
\centering
\caption{{Detailed Performance Breakdown on GenEval.} We report the fine-grained scores across six dimensions: Single Object, Two Object, Counting, Colors, Position, and Attribute Binding, with the overall performance shown in the last column.}
\label{tab:geneval_detailed}
\scriptsize
\resizebox{\columnwidth}{!}{
\begin{tabular}{l|l|cccccc|c}
\toprule
\textbf{Base Model} & \textbf{Method} & \textbf{Single Obj} & \textbf{Two Obj} & \textbf{Counting} & \textbf{Colors} & \textbf{Position} & \textbf{Attr Binding} & \textbf{Overall} \\ \midrule

\rowcolor{color1} \multicolumn{9}{c}{\textbf{Closed-Source Models}} \\ \midrule
Nano Banana & -- & -- & -- & -- & -- & -- & -- & 0.75 \\
GPT-Image 1 & -- & 0.99 & 0.92 & 0.85 & 0.92 & 0.75 & 0.61 & 0.84 \\
Seedream 4~\cite{seedream4} & -- & -- & -- & -- & -- & -- & -- & 0.84 \\ \midrule

\rowcolor{color2} \multicolumn{9}{c}{\textbf{Open-Source Models}} \\ \midrule
Bagel~\cite{bagel} & -- & -- & -- & -- & -- & -- & -- & 0.82 \\
Z-Image~\cite{z-image} & -- & 1.00 & 0.94 & 0.78 & 0.93 & 0.62 & 0.77 & 0.84 \\
Qwen-Image~\cite{qwen-image} & -- & 0.99 & 0.92 & 0.89 & 0.88 & 0.76 & 0.77 & 0.87 \\ \midrule

\rowcolor{color3} \multicolumn{9}{c}{\textbf{Inference-Time Scaling}} \\ \midrule
\multirow{7}{*}{Z-Image-Turbo} & Original & 1.00 & 0.92 & 0.73 & 0.87 & 0.49 & 0.61 & 0.77 \\
 & Rewrite & 1.00 & 0.89 & 0.69 & 0.89 & 0.60 & 0.56 & 0.77 \\
 & Promptist~\cite{promptist} & 1.00 & 0.91 & 0.70 & 0.86 & 0.51 & 0.59 & 0.76 \\
 & Search~\cite{ma2025inference} & 1.00 & 0.92 & 0.85 & 0.86 & 0.60 & 0.65 & 0.81 \\
 & Maestro~\cite{maestro} & 1.00 & 0.95 & 0.80 & 0.89 & 0.65 & 0.62 & 0.82 \\
 & CRAFT~\cite{craft} & 0.99 & 0.91 & 0.76 & 0.88 & 0.65 & 0.63 & 0.80 \\
 \rowcolor{color4} \cellcolor{white}& GEMS (Ours) & 0.99 & 0.92 & 0.91 & 0.88 & 0.83 & 0.65 & 0.86 \\ \midrule

\multirow{7}{*}{Qwen-Image-2512} & Original & 0.95 & 0.87 & 0.58 & 0.77 & 0.36 & 0.47 & 0.66 \\
 & Rewrite & 0.98 & 0.85 & 0.60 & 0.70 & 0.56 & 0.41 & 0.68 \\
 & Promptist~\cite{promptist} & 0.91 & 0.83 & 0.54 & 0.74 & 0.42 & 0.54 & 0.66 \\
 & Search~\cite{ma2025inference} & 1.00 & 0.94 & 0.61 & 0.81 & 0.63 & 0.47 & 0.74 \\
 & Maestro~\cite{maestro} & 1.00 & 0.93 & 0.70 & 0.73 & 0.62 & 0.58 & 0.76 \\
 & CRAFT~\cite{craft} & 1.00 & 0.90 & 0.74 & 0.82 & 0.70 & 0.58 & 0.79 \\
 \rowcolor{color4} \cellcolor{white}& GEMS (Ours) & 0.98 & 0.94 & 0.73 & 0.84 & 0.72 & 0.54 & 0.79 \\ \bottomrule
\end{tabular}
}
\end{table*}

\begin{table*}[htbp]
\centering
\caption{{Detailed Performance Breakdown on GenEval2.} We report the fine-grained scores across five dimensions: Object, Attribute, Count, Position, and Verb.}
\label{tab:geneval2_detailed}
\scriptsize
\begin{tabular}{l|l|ccccc}
\toprule
\textbf{Base Model} & \textbf{Method} & \textbf{Object} & \textbf{Attribute} & \textbf{Count} & \textbf{Position} & \textbf{Verb}  \\ \midrule
    
\rowcolor{color1} \multicolumn{7}{c}{\textbf{Closed-Source Models}} \\ \midrule
Nano Banana & -- & 99.0&91.4&70.1&70.2&86.7  \\
Nano Banana 2 & -- & 97.0&95.1&76.4&83.7&56.1  \\\midrule

\rowcolor{color2} \multicolumn{7}{c}{\textbf{Open-Source Models}} \\ \midrule
Bagel~\cite{bagel} & -- & 92.9&75.9&55.6&50.6&57.8   \\
Qwen-Image~\cite{qwen-image} & -- & 99.1&85.6&70.3&60.2&71.1\\ \midrule

\rowcolor{color3} \multicolumn{7}{c}{\textbf{Inference-Time Scaling}} \\ \midrule
\multirow{7}{*}{Z-Image-Turbo} & Original & 97.8 & 75.4 & 62.8 & 62.2 & 18.6   \\
 & Rewrite & 97.2 & 86.8 & 64.9 & 73.2 & 27.6  \\
 & Promptist~\cite{promptist} & 98.4 & 78.2 & 63.1 & 65.6 & 23.1 \\
 & Search~\cite{ma2025inference} & 98.4 & 78.1 & 73.8 & 67.1 & 22.9   \\
 & Maestro~\cite{maestro} & 96.6 & 87.1 & 67.2 & 75.8 & 36.0   \\
 & CRAFT~\cite{craft} & 98.1 & 93.1 & 78.0 & 86.7 & 45.6  \\
  \rowcolor{color4} \cellcolor{white}&GEMS (Ours) & 97.9 & 92.4 & 79.0 & 85.4& 41.1   \\ \midrule

\multirow{7}{*}{Qwen-Image-2512} & Original &97.9  &65.2  & 66.9 &  58.1&22.1    \\
 & Rewrite &  97.3&81.1  &64.0  & 73.8 &30.3   \\
 & Promptist~\cite{promptist} & 97.3 & 64.4 & 63.0 & 59.4 & 28.6   \\
 & Search~\cite{ma2025inference} & 98.3 &71.1  &73.9  &66.1  &28.2    \\
 & Maestro~\cite{maestro} & 97.3 & 84.9 &68.0  & 80.9 &31.0  \\
 & CRAFT~\cite{craft} & 97.9 & 91.3 & 79.0 & 89.9 & 53.7   \\
  \rowcolor{color4} \cellcolor{white}&GEMS (Ours) & 96.9 &91.4  &81.3  & 91.4 &45.5    \\ \bottomrule
\end{tabular}
\end{table*}

\begin{table*}[htbp]
\centering
\caption{{Detailed Performance Breakdown on DPG-Bench.} We report the fine-grained scores across five dimensions: Global, Entity, Attribute, Relation, and Other, with the overall performance shown in the last column.}
\label{tab:dpg_bench_detailed}
\scriptsize
\begin{tabular}{l|l|ccccc|c}
\toprule
\textbf{Base Model} & \textbf{Method} & \textbf{Global} & \textbf{Entity} & \textbf{Attribute} & \textbf{Relation} & \textbf{Other} & \textbf{Overall} \\ \midrule

\rowcolor{color1} \multicolumn{8}{c}{\textbf{Closed-Source Models}} \\ \midrule
Nano Banana & -- & -- & -- & -- & -- & -- & 85.23 \\
GPT-Image 1 & -- & 88.89 & 88.94 & 89.84 & 92.63 & 90.96 & 85.15 \\
Seedream 4~\cite{seedream4} & -- & -- & -- & -- & -- & -- & 88.25 \\ \midrule

\rowcolor{color2} \multicolumn{8}{c}{\textbf{Open-Source Models}} \\ \midrule
Bagel~\cite{bagel} & -- & -- & -- & -- & -- & -- & 85.10 \\
Z-Image~\cite{z-image} & -- & 93.39 & 91.22 & 93.16 & 92.22 & 91.52 & 88.14 \\
Qwen-Image~\cite{qwen-image} & -- & 91.32 & 91.56 & 92.02 & 94.31 & 92.73 & 88.32 \\ \midrule

\rowcolor{color3} \multicolumn{8}{c}{\textbf{Inference-Time Scaling}} \\ \midrule
\multirow{7}{*}{Z-Image-Turbo} & Original & 90.56 & 90.40 & 90.70 & 91.39 & 90.58 & 85.08 \\
 & Rewrite & 85.23 & 88.43 & 90.23 & 90.18 & 92.01 & 84.48 \\
 & Promptist~\cite{promptist} & 70.78 & 72.87 & 72.02 & 73.12 & 73.56 & 65.69 \\
 & Search~\cite{ma2025inference} & 91.34 & 89.60 & 90.62 & 92.51 & 91.06 & 85.47 \\
 & Maestro~\cite{maestro} & 87.13 & 91.24 & 90.66 & 91.15 & 91.25 & 85.29 \\
 & CRAFT~\cite{craft} & 93.15 & 91.41 & 89.16 & 91.80 & 90.38 & 85.29 \\
  \rowcolor{color4} \cellcolor{white}&GEMS (Ours) & 90.82 & 89.97 & 90.48 & 93.11 & 90.61 & 86.01 \\ \midrule

\multirow{7}{*}{Qwen-Image-2512} & Original & 88.36 & 90.70 & 90.33 & 91.57 & 89.88 & 84.69 \\
 & Rewrite & 91.66 & 89.19 & 88.42 & 92.07 & 88.99 & 83.49 \\
 & Promptist~\cite{promptist} & 72.92 & 71.97 & 71.26 & 71.87 & 73.38 & 64.02 \\
 & Search~\cite{ma2025inference} & 92.72 & 90.31 & 89.34 & 93.74 & 89.81 & 86.17 \\
 & Maestro~\cite{maestro} & 92.11 & 88.79 & 89.24 & 89.35 & 91.62 & 84.07 \\
 & CRAFT~\cite{craft} & 91.34 & 92.07 & 88.56 & 92.74 & 89.74 & 85.87 \\
 \rowcolor{color4} \cellcolor{white} & GEMS (Ours) & 87.39 & 91.81 & 90.00 & 93.45 & 92.37 & 85.59 \\ \bottomrule
\end{tabular}
\end{table*}

\begin{table*}[htbp]
\centering
\caption{{Detailed Performance Breakdown on OneIG-EN.} We report the fine-grained scores across five dimensions: Alignment, Text, Reasoning, Style, and Diversity, with the overall performance shown in the last column.}
\label{tab:oneig_en_detailed}
\scriptsize
\begin{tabular}{l|l|ccccc|c}
\toprule
\textbf{Base Model} & \textbf{Method} & \textbf{Alignment} & \textbf{Text} & \textbf{Reasoning} & \textbf{Style} & \textbf{Diversity} & \textbf{Overall} \\ \midrule

\rowcolor{color1} \multicolumn{8}{c}{\textbf{Closed-Source Models}} \\ \midrule
Nano Banana & -- & 0.878 & 0.894 & 0.346 & 0.450 & 0.182 & 0.550 \\
GPT-Image 1 & -- & 0.851 & 0.857 & 0.345 & 0.462 & 0.151 & 0.533 \\
Seedream 4~\cite{seedream4} & -- & 0.894 & 0.981 & 0.352 & 0.458 & 0.197 & 0.576 \\ \midrule

\rowcolor{color2} \multicolumn{8}{c}{\textbf{Open-Source Models}} \\ \midrule
Bagel~\cite{bagel} & -- & 0.769 & 0.244 & 0.173 & 0.367 & 0.251 & 0.361 \\
Z-Image~\cite{z-image} & -- & 0.881 & 0.987 & 0.280 & 0.387 & 0.194 & 0.546 \\
Qwen-Image~\cite{qwen-image} & -- & 0.882 & 0.891 & 0.306 & 0.418 & 0.197 & 0.539 \\ \midrule

\rowcolor{color3} \multicolumn{8}{c}{\textbf{Inference-Time Scaling}} \\ \midrule
\multirow{7}{*}{Z-Image-Turbo} & Original & 0.836 & 0.987 & 0.302 & 0.370 & 0.137 & 0.526 \\
 & Rewrite & 0.835 & 0.839 & 0.277 & 0.433 & 0.215 & 0.520 \\
 & Promptist~\cite{promptist} & 0.667 & 0.266 & 0.264 & 0.308 & 0.161 & 0.333 \\
 & Search~\cite{ma2025inference} & 0.845 & 0.990 & 0.306 & 0.376 & 0.134 & 0.530 \\
 & Maestro~\cite{maestro} & 0.843 & 0.964 & 0.294 & 0.427 & 0.212 & 0.548 \\
 & CRAFT~\cite{craft} & 0.863 & 0.980 & 0.309 & 0.460 & 0.300 & 0.582 \\
  \rowcolor{color4} \cellcolor{white}&GEMS (Ours) & 0.869 & 0.983 & 0.312 & 0.451 & 0.232 & 0.569 \\ \midrule

\multirow{7}{*}{Qwen-Image-2512} & Original & 0.830 & 0.910 & 0.235 & 0.233 & 0.225 & 0.487 \\
 & Rewrite & 0.828 & 0.752 & 0.249 & 0.265 & 0.231 & 0.465 \\
 & Promptist~\cite{promptist} & 0.650 & 0.166 & 0.208 & 0.173 & 0.242 & 0.288 \\
 & Search~\cite{ma2025inference} & 0.848 & 0.944 & 0.248 & 0.254 & 0.218 & 0.502 \\
 & Maestro~\cite{maestro} & 0.836 & 0.921 & 0.264 & 0.283 & 0.166 & 0.494 \\
 & CRAFT~\cite{craft} & 0.858 & 0.944 & 0.289 & 0.331 & 0.241 & 0.533 \\
 \rowcolor{color4} \cellcolor{white} & GEMS (Ours) & 0.859 & 0.972 & 0.282 & 0.333 & 0.262 & 0.542 \\ \bottomrule
\end{tabular}
\end{table*}

\begin{table*}[htbp]
\centering
\caption{{Detailed Performance Breakdown on OneIG-ZH.} We report the fine-grained scores across five dimensions: Alignment, Text, Reasoning, Style, and Diversity, with the overall performance shown in the last column.}
\label{tab:oneig_zh_detailed}
\scriptsize
\begin{tabular}{l|l|ccccc|c}
\toprule
\textbf{Base Model} & \textbf{Method} & \textbf{Alignment} & \textbf{Text} & \textbf{Reasoning} & \textbf{Style} & \textbf{Diversity} & \textbf{Overall} \\ \midrule

\rowcolor{color1} \multicolumn{8}{c}{\textbf{Closed-Source Models}} \\ \midrule
Nano Banana & -- & 0.825 & 0.276 & 0.298 & 0.427 & 0.198 & 0.405 \\
GPT-Image 1 & -- & 0.812 & 0.650 & 0.300 & 0.449 & 0.159 & 0.474 \\
Seedream 4~\cite{seedream4} & -- & 0.847 & 0.982 & 0.286 & 0.443 & 0.206 & 0.553 \\ \midrule

\rowcolor{color2} \multicolumn{8}{c}{\textbf{Open-Source Models}} \\ \midrule
Bagel~\cite{bagel} & -- & 0.672 & 0.365 & 0.186 & 0.357 & 0.268 & 0.370 \\
Z-Image~\cite{z-image} & -- & 0.793 & 0.988 & 0.266 & 0.386 & 0.243 & 0.535 \\
Qwen-Image~\cite{qwen-image} & -- & 0.825 & 0.963 & 0.267 & 0.405 & 0.279 & 0.548 \\ \midrule

\rowcolor{color3} \multicolumn{8}{c}{\textbf{Inference-Time Scaling}} \\ \midrule
\multirow{7}{*}{Z-Image-Turbo} & Original & 0.773 & 0.958 & 0.275 & 0.362 & 0.138 & 0.501 \\
 & Rewrite & 0.788 & 0.645 & 0.269 & 0.424 & 0.217 & 0.469 \\
 & Promptist~\cite{promptist} & 0.377 & 0.510 & 0.250 & 0.292 & 0.207 & 0.327 \\
 & Search~\cite{ma2025inference} & 0.784 & 0.964 & 0.280 & 0.365 & 0.135 & 0.506 \\
 & Maestro~\cite{maestro} & 0.792 & 0.896 & 0.281 & 0.405 & 0.220 & 0.519 \\
 & CRAFT~\cite{craft} & 0.805 & 0.863 & 0.288 & 0.446 & 0.306 & 0.542 \\
  \rowcolor{color4} \cellcolor{white}&GEMS (Ours) & 0.800 & 0.978 & 0.294 & 0.448 & 0.240 & 0.552 \\ \midrule

\multirow{7}{*}{Qwen-Image-2512} & Original & 0.784 & 0.966 & 0.243 & 0.226 & 0.227 & 0.489 \\
 & Rewrite & 0.775 & 0.658 & 0.250 & 0.263 & 0.231 & 0.435 \\
 & Promptist~\cite{promptist} & 0.405 & 0.432 & 0.216 & 0.189 & 0.293 & 0.307 \\
 & Search~\cite{ma2025inference} & 0.796 & 0.976 & 0.251 & 0.244 & 0.217 & 0.497 \\
 & Maestro~\cite{maestro} & 0.785 & 0.904 & 0.264 & 0.258 & 0.235 & 0.489 \\
 & CRAFT~\cite{craft} & 0.801 & 0.875 & 0.274 & 0.326 & 0.313 & 0.518 \\
 \rowcolor{color4} \cellcolor{white}& GEMS (Ours) & 0.810 & 0.979 & 0.279 & 0.325 & 0.267 & 0.532 \\ \bottomrule
\end{tabular}
\end{table*}

\begin{table*}[htbp]
\centering
\caption{{Detailed Performance Breakdown on WISE.} We report the fine-grained scores across six dimensions: Cultural, Time, Space, Biology, Physics, and Chemistry, with the overall performance shown in the last column.}
\label{tab:wise_detailed}
\scriptsize
\begin{tabular}{l|l|cccccc|c}
\toprule
\textbf{Base Model} & \textbf{Method} & \textbf{Cultural} & \textbf{Time} & \textbf{Space} & \textbf{Biology} & \textbf{Physics} & \textbf{Chemistry} & \textbf{Overall} \\ \midrule

\rowcolor{color1} \multicolumn{9}{c}{\textbf{Closed-Source Models}} \\ \midrule
GPT-Image 1 & -- & 0.81 & 0.71 & 0.89 & 0.83 & 0.79 & 0.74 & 0.80 \\
Seedream 4~\cite{seedream4} & -- & 0.78 & 0.73 & 0.85 & 0.79 & 0.84 & 0.67 & 0.78 \\ \midrule

\rowcolor{color2} \multicolumn{9}{c}{\textbf{Open-Source Models}} \\ \midrule
Bagel~\cite{bagel} & -- & 0.76 & 0.69 & 0.75 & 0.65 & 0.75 & 0.58 & 0.70 \\
Qwen-Image~\cite{qwen-image} & -- & 0.62 & 0.63 & 0.77 & 0.57 & 0.75 & 0.40 & 0.62 \\ \midrule

\rowcolor{color3} \multicolumn{9}{c}{\textbf{Inference-Time Scaling}} \\ \midrule
\multirow{7}{*}{Z-Image-Turbo} & Original & 0.57 & 0.58 & 0.71 & 0.50 & 0.60 & 0.39 & 0.57 \\
 & Rewrite & 0.88 & 0.78 & 0.89 & 0.80 & 0.81 & 0.81 & 0.84 \\
 & Promptist~\cite{promptist} & 0.54 & 0.53 & 0.68 & 0.45 & 0.58 & 0.39 & 0.54 \\
 & Search~\cite{ma2025inference} & 0.60 & 0.59 & 0.78 & 0.56 & 0.66 & 0.44 & 0.61 \\
 & Maestro~\cite{maestro} & 0.88 & 0.78 & 0.89 & 0.81 & 0.85 & 0.83 & 0.85 \\
 & CRAFT~\cite{craft} & 0.83	& 0.72 & 0.77 & 0.76 & 0.76 & 0.72 & 0.78 \\
  \rowcolor{color4} \cellcolor{white}&GEMS (Ours) & 0.81 & 0.76 & 0.86 & 0.82 & 0.81 & 0.79 & 0.81 \\ \midrule

\multirow{7}{*}{Qwen-Image-2512} & Original & 0.60 & 0.61 & 0.68 & 0.51 & 0.66 & 0.37 & 0.59 \\
 & Rewrite & 0.86 & 0.76 & 0.86 & 0.78 & 0.80 & 0.71 & 0.81 \\
 & Promptist~\cite{promptist} & 0.56 & 0.59 & 0.69 & 0.52 & 0.66 & 0.37 & 0.57 \\
 & Search~\cite{ma2025inference} & 0.67 & 0.64 & 0.81 & 0.64 & 0.73 & 0.49 & 0.67 \\
 & Maestro~\cite{maestro} & 0.85 & 0.79 & 0.87 & 0.85 & 0.86 & 0.77 & 0.84 \\
 & CRAFT~\cite{craft} & 0.84 & 0.73 & 0.81 & 0.75 & 0.76 & 0.76 & 0.79 \\
  \rowcolor{color4} \cellcolor{white}&GEMS (Ours) & 0.81 & 0.77 & 0.86 & 0.79 & 0.78 & 0.77 & 0.80 \\ \bottomrule
\end{tabular}
\end{table*}

\begin{table*}[htbp]
\centering
\caption{{Detailed Performance Breakdown on SpatialGenEval.} We report the fine-grained scores across ten dimensions: Object, Attribute, Position, Orientation, Layout, Comparison, Proximity, Occlusion, Motion, and Causal, with the overall performance shown in the last column.}
\label{tab:spatialgeneval_detailed}
\scriptsize
\resizebox{\columnwidth}{!}{%
\begin{tabular}{l|l|cccccccccc|c}
\toprule
\textbf{Base Model} & \textbf{Method} & \textbf{Object} & \textbf{Attribute} & \textbf{Position} & \textbf{Orientation} & \textbf{Layout} & \textbf{Comparison}& \textbf{Proximity}& \textbf{Occlusion}& \textbf{Motion}& \textbf{Causal} & \textbf{Overall} \\ \midrule

\rowcolor{color1} \multicolumn{13}{c}{\textbf{Closed-Source Models}} \\ \midrule
Nano Banana & -- & 58.5& 75.3 &55.5& 58.9 &70.9 &31.8 &68.7 &33.5& 81.4 &82.2&61.7 \\
GPT-Image 1 & -- & 56.3 &74.1& 53.3& 58.9 &70.4 &31.4 &66.8& 30.2 &80.9 &82.2&60.5 \\
Seedream 4~\cite{seedream4} & -- & 59.9& 80.2& 57.2 &58.9 &70.1 &32.1 &68.3& 33.8 &83.0& 83.8 &62.7\\\midrule

\rowcolor{color2} \multicolumn{13}{c}{\textbf{Open-Source Models}} \\ \midrule
Bagel~\cite{bagel} & -- & 55.3&73.7 &51.2& 54.0& 62.9& 28.6& 64.1& 29.0 &74.4& 76.7&57.0 \\
Qwen-Image~\cite{qwen-image} & -- & 61.0& 77.2 &55.6 &56.7& 69.7 &28.6 &67.7 &30.8 &78.1 &80.2&60.6 \\ \midrule

\rowcolor{color3} \multicolumn{13}{c}{\textbf{Inference-Time Scaling}} \\ \midrule
\multirow{7}{*}{Z-Image-Turbo} & Original &56.5 & 74.7 & 53.2 & 56.0 & 67.5 & 29.9 & 66.3 & 29.6 & 75.8 & 77.1 & 58.7 \\
 & Rewrite & 56.6 & 74.2 & 53.0 & 57.9 & 67.8 & 28.9 & 67.0 & 30.6 & 79.7 & 79.8 & 59.5 \\
 & Promptist~\cite{promptist} & 37.7 & 65.9 & 47.6 & 52.6 & 56.3 & 22.4 & 52.9 & 24.3 & 61.2 & 62.1 & 48.3 \\
 & Search~\cite{ma2025inference} & 56.7 & 75.8 & 55.5 & 57.0 & 67.9 & 28.9 & 66.1 & 29.8 & 76.3 & 79.3 & 59.3 \\
 & Maestro~\cite{maestro} & 59.0 & 74.7 & 53.9 & 58.9 & 68.3 & 30.5 & 67.2 & 30.8 & 79.4 & 81.1 & 60.4 \\
 & CRAFT~\cite{craft} & 56.3 & 75.9 & 54.0 & 57.1 & 69.7 & 31.5 & 67.3 & 30.8 & 82.0 & 81.2 & 60.6 \\
  \rowcolor{color4} \cellcolor{white}&GEMS (Ours) & 58.9 & 75.9 & 55.4 & 57.4 & 71.1 & 32.0 & 69.0 & 31.1 & 81.1 & 82.1 & 61.4 \\ \midrule
\multirow{7}{*}{Qwen-Image-2512} & Original & 56.8 & 76.2 & 56.8 & 58.5 & 67.6 & 28.5 & 68.0 & 31.9 & 77.2 & 78.6 & 60.0 \\
 & Rewrite &57.6 & 75.6 & 54.3 & 58.1 & 68.0 & 28.1 & 69.4 & 32.2 & 79.8 & 80.3 & 60.4 \\
 & Promptist~\cite{promptist} & 39.3 & 65.1 & 47.6 & 50.8 & 55.4 & 21.9 & 55.8 & 25.0 & 62.9 & 62.9 & 48.7 \\
 & Search~\cite{ma2025inference} &56.3 & 76.5 & 55.5 & 60.2 & 68.6 & 30.1 & 69.6 & 30.6 & 79.3 & 79.6 & 60.6 \\
 & Maestro~\cite{maestro} & 56.9 & 75.5 & 57.0 & 59.3 & 68.7 & 30.1 & 69.4 & 32.6 & 79.8 & 81.1 & 61.0  \\
 & CRAFT~\cite{craft} & 59.3 & 75.8 & 54.4 & 60.5 & 68.5 & 31.6 & 68.0 & 32.6 & 82.9 & 82.7 & 61.6 \\
  \rowcolor{color4} \cellcolor{white}&GEMS (Ours) &61.3 & 76.3 & 57.7 & 58.9 & 70.7 & 31.1 & 68.8 & 32.8 & 81.5 & 82.0 & 62.1 \\ \bottomrule
\end{tabular}
}
\end{table*}

\begin{table*}[htbp]
\centering
\caption{{Detailed Performance Breakdown on CREA.} We report the fine-grained scores across six dimensions: Originality, Expressiveness, Aesthetic, Technical, Unexpected, and Interpretability, with the overall performance shown in the last column.}
\label{tab:crea_detailed}
\scriptsize
\resizebox{\columnwidth}{!}{%
\begin{tabular}{l|l|cccccc|c}
\toprule
\textbf{Base Model} & \textbf{Method} & \textbf{Originality} & \textbf{Expressiveness} & \textbf{Aesthetic} & \textbf{Technical } & \textbf{Unexpected} & \textbf{Interpretability} & \textbf{Overall} \\ \midrule
\rowcolor{color3} \multicolumn{9}{c}{\textbf{Inference-Time Scaling}} \\ \midrule
\multirow{7}{*}{Z-Image-Turbo} & Original &1.62  &1.57  &   2.92&   3.19&  1.37 &  1.17 &11.84 \\ 
 & Rewrite &  2.39&  2.63&   3.72 &3.80  & 2.22 &1.99&16.75 \\
 & Promptist~\cite{promptist} & 1.85 & 2.02 & 3.36 & 3.76 & 1.57 & 1.45 & 14.00 \\
 & Search~\cite{ma2025inference} & 1.87 & 1.88& 3.07 & 3.33 & 1.56 & 1.28 & 12.98 \\
 & Maestro~\cite{maestro} & 2.32 & 2.46 & 3.66 & 3.76 & 1.92 & 1.68 &15.81 \\
 & CRAFT~\cite{craft} & 1.98 & 2.01 & 3.20 & 3.44 & 1.57 & 1.43 & 13.63 \\
  \rowcolor{color4} \cellcolor{white}&GEMS (Ours) & 3.36 &3.48 & 4.36 & 4.45 & 3.57 &3.33& 22.55 \\ \midrule
\multirow{7}{*}{Qwen-Image-2512} & Original &2.46&2.81&3.67&3.89&2.36&2.28  &17.45\\
 & Rewrite & 2.63 &2.83  &  3.96& 4.07 & 2.38 &2.25  & 18.12 \\
 & Promptist~\cite{promptist} & 2.54 & 2.78 & 3.95 & 4.41 & 2.32 & 2.35 & 18.35 \\
 & Search~\cite{ma2025inference} &2.69  &2.95  & 3.86 &4.07  &  2.58&  2.56& 18.71 \\
 & Maestro~\cite{maestro} & 2.59 & 2.76 & 3.93 & 4.05 & 2.34 &2.12  &17.78  \\
 & CRAFT~\cite{craft} & 2.30 & 2.42 & 3.57 & 3.74 & 1.88 & 1.70 & 15.61 \\
  \rowcolor{color4} \cellcolor{white}&GEMS (Ours) & 3.52 & 3.87 & 4.59 & 4.65 & 3.69 &3.68  & 24.01 \\ \bottomrule
\end{tabular}
}
\end{table*}

\section{Qualitative Results}
\begin{figure}[htbp]
    \centering
    \begin{minipage}{0.45\textwidth}
        \flushright
        \includegraphics[width=0.7\textwidth]{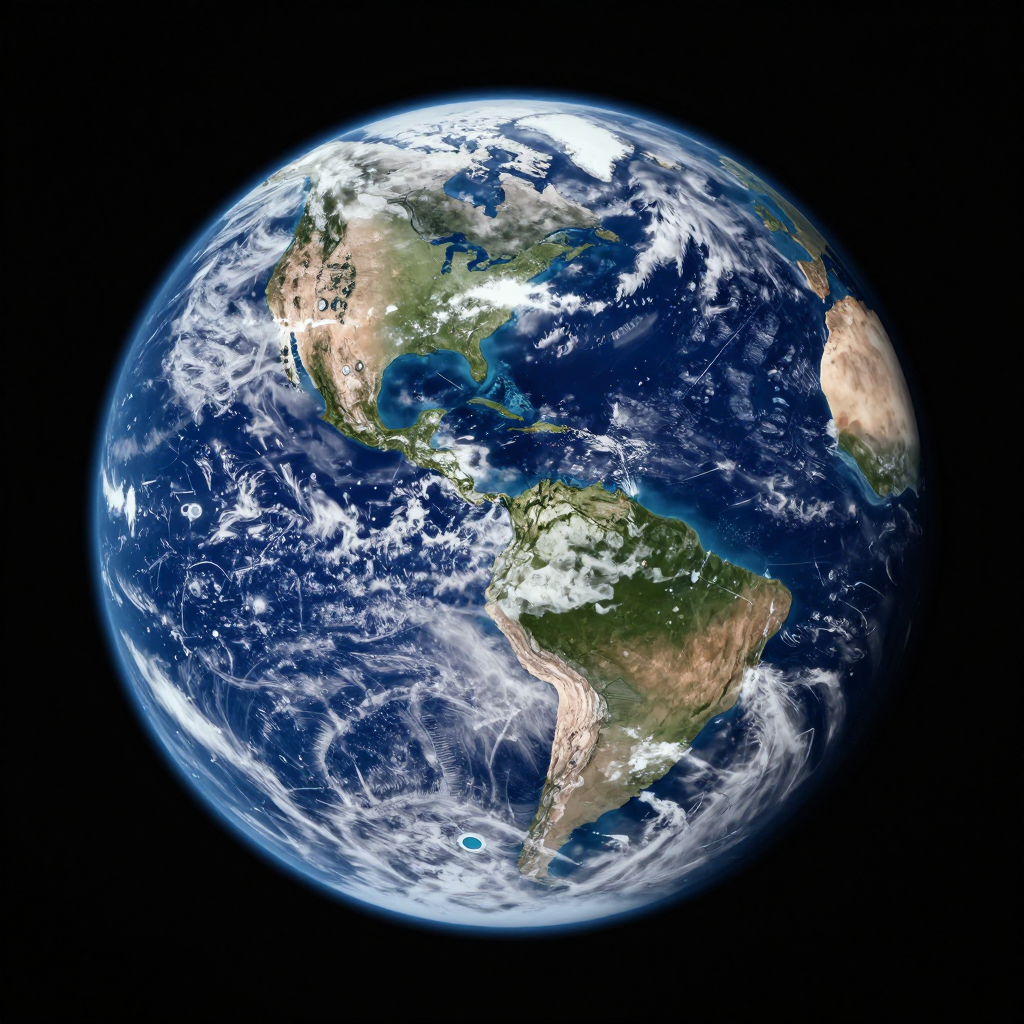}
    \end{minipage}
    \hfill
    \begin{minipage}{0.45\textwidth}
        \flushleft
        \includegraphics[width=0.7\textwidth]{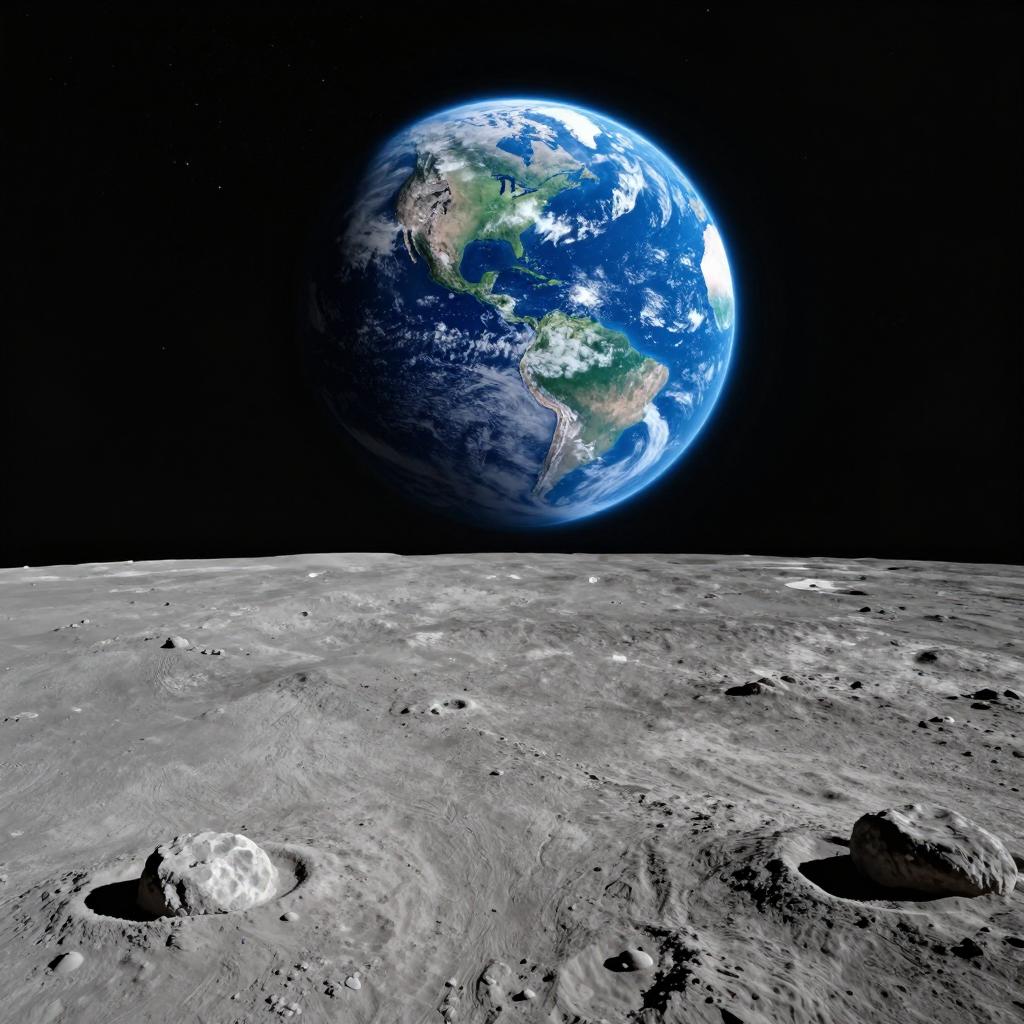}
    \end{minipage}
    \caption{Baseline vs GEMS for prompt: \textit{a view of the Earth from the moon.}}
    \label{fig:qualitative_1}
\end{figure}

\begin{figure}[htbp]
    \centering
    \begin{minipage}{0.45\textwidth}
        \flushright
        \includegraphics[width=0.7\textwidth]{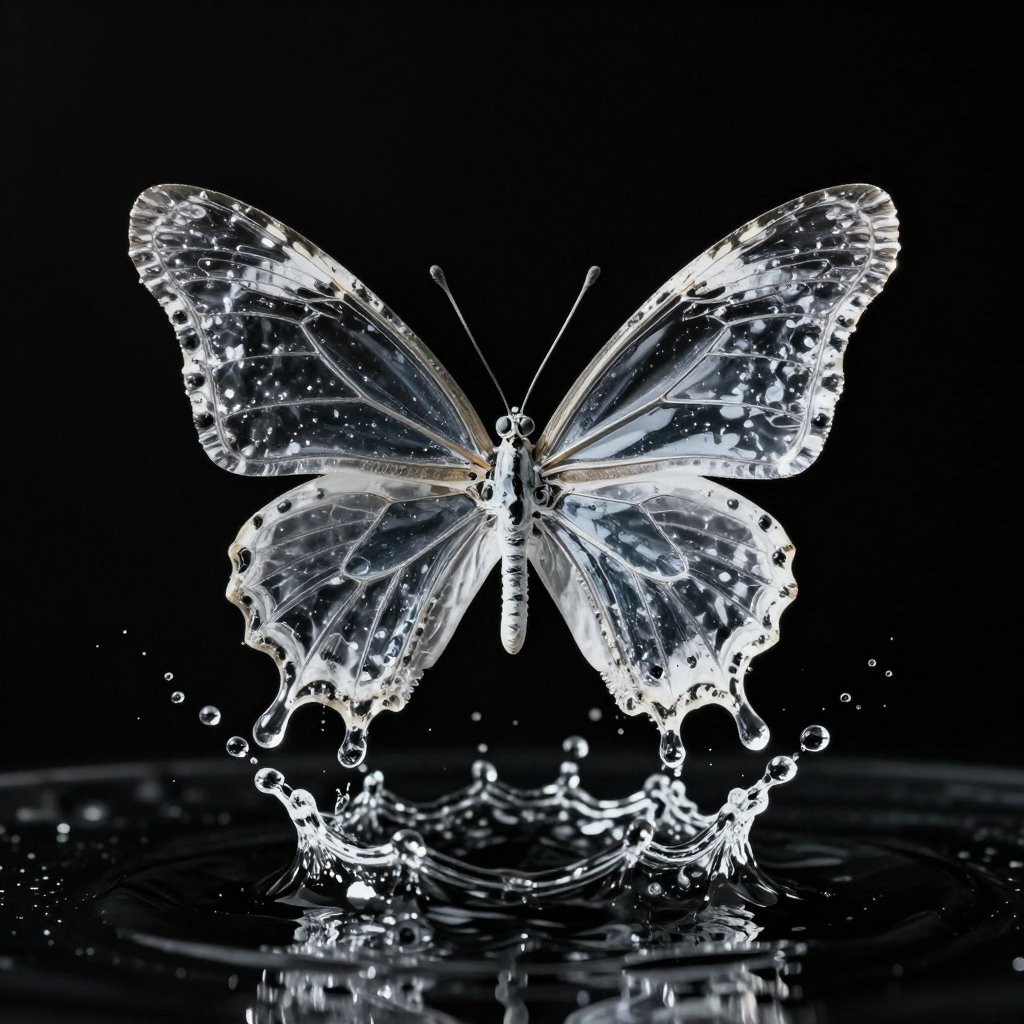}
    \end{minipage}
    \hfill
    \begin{minipage}{0.45\textwidth}
        \flushleft
        \includegraphics[width=0.7\textwidth]{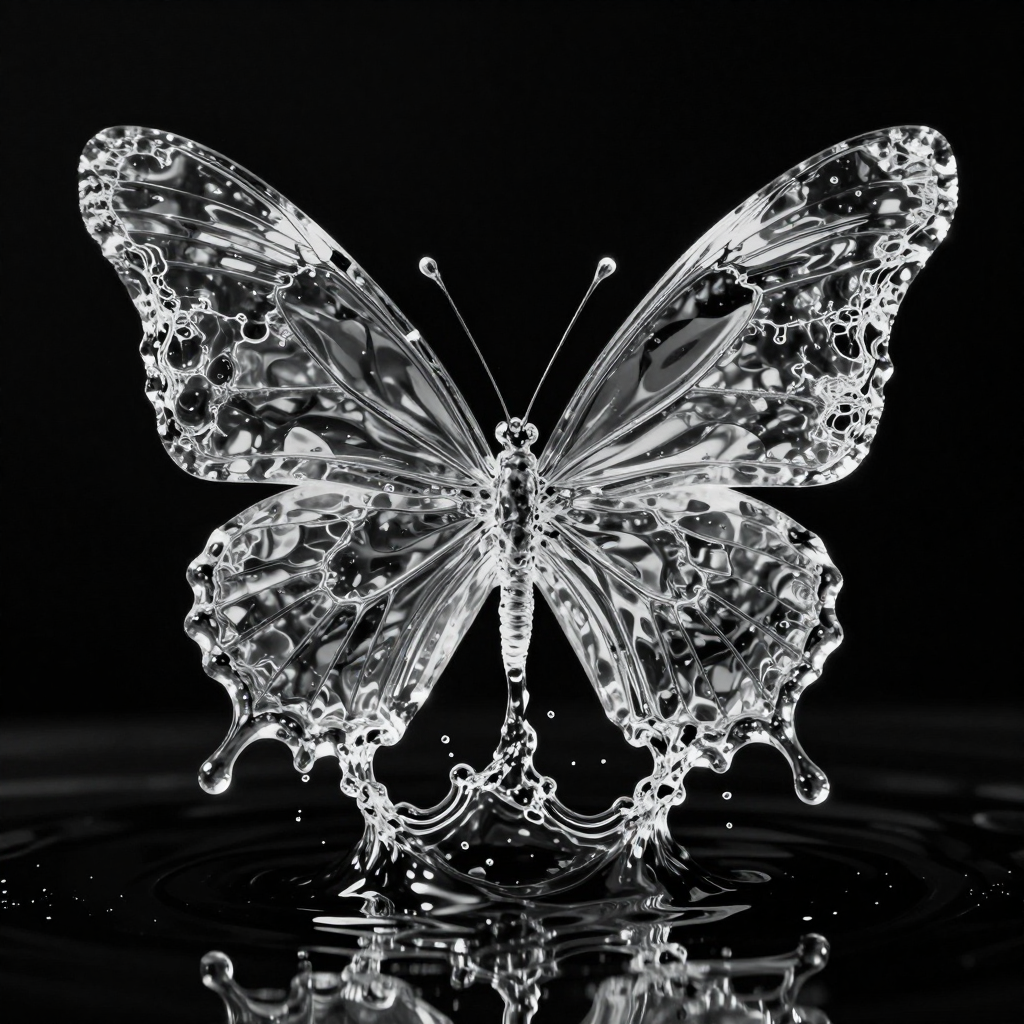}
    \end{minipage}
    \caption{Baseline vs GEMS for prompt: \textit{A high-speed photography shot of clear water being splashed upwards from a black surface. In mid-air, the splashing water droplets perfectly form the shape of a detailed, symmetrical butterfly. The wings are made entirely of liquid water ripples. The background is completely pitch black. The water butterfly is physically connected to the splash below.}}
    \label{fig:qualitative_2}
\end{figure}

In addition to Figure~\ref{fig:skill_demo}, we further present comparisons of the results generated by GEMS and the baseline with same prompt, as shown in Figure~\ref{fig:qualitative_1} through Figure~\ref{fig:qualitative_2} 

\section{Limitations and Future Work}

First, despite utilizing the lightweight and distilled Z-Image-Turbo, the iterative nature of Agent Loop still results in noticeable inference latency. Future work will focus on optimizing the workflow design to minimize computational overhead and improve overall efficiency.

Second, the current system relies on predefined workflows to coordinate agent collaboration and module interactions. We plan to investigate higher levels of agent autonomy in the future, such as providing tool interfaces that allow models to autonomously manage memory and skills.

Third, while the current system is primarily designed for image generation, GEMS has the potential to be extended to other multimodal tasks. Future research will explore its application in more complex domains such as video generation.

Finally, current Z-Image-Turbo and Qwen-Image-2512 do not support image editing. Future research could leverage more versatile models to evolve GEMS into a comprehensive Agent-Native system that integrates reasoning, generation, and editing in a unified intelligence loop.

\end{document}